\documentclass[preprint,12pt,authoryear]{elsarticle}

\usepackage{amsmath,amsfonts}
\usepackage[caption=false,font=normalsize,labelfont=sf,textfont=sf]{subfig}
\usepackage{xcolor}
\usepackage[ruled,linesnumbered]{algorithm2e} % IEEE algorithmic, algorithm after algorithm2e

\usepackage{amssymb}
\usepackage{url}
\newtheorem{definition}{Definition}

\newtheorem{problem}{Problem}

\newlength\savewidth\newcommand\shline{\noalign{\global\savewidth\arrayrulewidth
		\global\arrayrulewidth 1pt}\hline\noalign{\global\arrayrulewidth\savewidth}}

\usepackage{booktabs}
\usepackage{multirow}

\usepackage{array}
\usepackage{diagbox}
\usepackage{wrapfig}

\journal{Neural Networks}

\begin{document}

\begin{frontmatter}



\title{Learning Dynamic Graph Representations through Timespan View Contrasts}


% \author{}

% \affiliation{organization={},%Department and Organization

% Yiming Xu, Zhen Peng, Bin Shi, Xu Hua, Bo Dong, Qinghua Zheng

\author[inst1]{Yiming Xu}\ead{xym0924@stu.xjtu.edu.cn}

\affiliation[inst1]{organization={School of Computer Science and Technology},%Department and Organization
            addressline={Xi'an Jiaotong University}, 
            %city={},
            %postcode={}, 
            %state={},
            country={P.R. China}}
\author[inst1]{Zhen Peng\corref{cor1}}\ead{zhenpeng27@outlook.com}
\author[inst1]{Bin Shi}\ead{shibin@xjtu.edu.cn}
\author[inst1]{Xu Hua}\ead{huaxu@stu.xjtu.edu.cn}
\author[inst2]{Bo Dong}\ead{dong.bo@xjtu.edu.cn}
% \author[inst1]{Qinghua Zheng}\ead{qhzheng@mail.xjtu.edu.cn}

\affiliation[inst2]{organization={School of Distance Education},%Department and Organization
            addressline={Xi'an Jiaotong University}, 
            %city={},
            %postcode={}, 
            %state={},
            country={P.R. China}}
\cortext[cor1]{Corresponding author.}

\begin{abstract}
The rich information underlying graphs has inspired further investigation of unsupervised graph representation. Existing studies mainly depend on node features and topological properties within static graphs to create self-supervised signals, neglecting the temporal components carried by real-world graph data, such as timestamps of edges. To overcome this limitation, this paper explores how to model temporal evolution on dynamic graphs elegantly. Specifically, we introduce a new inductive bias, namely temporal translation invariance, which illustrates the tendency of the identical node to keep similar labels across different timespans. Based on this assumption, we develop a dynamic graph representation framework CLDG that encourages the node to maintain locally consistent temporal translation invariance through contrastive learning on different timespans. Except for standard CLDG which only considers explicit topological links, our further proposed CLDG++ additionally employs graph diffusion to uncover global contextual correlations between nodes, and designs a multi-scale contrastive learning objective composed of local-local, local-global, and global-global contrasts to enhance representation capabilities. Interestingly, by measuring the consistency between different timespans to shape anomaly indicators, CLDG and CLDG++ are seamlessly integrated with the task of spotting anomalies on dynamic graphs, which has broad applications in many high-impact domains, such as finance, cybersecurity, and healthcare. Experiments demonstrate that CLDG and CLDG++ both exhibit desirable performance in downstream tasks including node classification and dynamic graph anomaly detection. Moreover, CLDG significantly reduces time and space complexity by implicitly exploiting temporal cues instead of complicated sequence models. The code and data are available at \url{https://github.com/yimingxu24/CLDG}.
\end{abstract}


% \begin{highlights}
% % \item Research highlight 1
% % \item Research highlight 2
% \item We propose a new inductive bias, temporal translation invariance, on dynamic graphs.

% \item We introduce a simpler and lighter dynamic graph representation learning method.

% \item We present a new discriminator and generalize our models to anomaly detection tasks.

% \item Experiments on 7 datasets and 2 tasks demonstrate the effectiveness of our model.
% \end{highlights}


\end{frontmatter}


\section{Introduction}
\label{sec:introduction}
Graph data is prevalent in various real-world domains. In recent years, research on high-quality representation learning for graph-structured data has attracted much attention~\citep{xu2025out,xu2025court}. Traditional representation learning methods often struggle to handle the complexity of graph-structured data effectively. In contrast, graph neural networks (GNNs) provide a novel perspective for modeling and analyzing graph data. GNNs have demonstrated remarkable performance in a wide range of applications, including recommender systems~\citep{wu2020graph,zhou2024node}, combinatorial optimization~\citep{li2018combinatorial, cappart2021combinatorial}, trafﬁc prediction~\citep{derrow2021eta,fan2024rgdan}, and risk management~\citep{zheng2023survey, song2024enhancing,xu2025ted}. \looseness=-1

Despite the empirical success, most existing GNN methods typically focus on static graphs and heavily rely on supervised learning paradigms. It is undeniable that real-world graph data inherently carries temporal information, such as timestamps of edges. The loss of information caused by ignoring these temporal components would lead to performance corruption. In addition, acquiring sufficient and high-quality labels in graph data necessitates rich domain knowledge and experience, which leads to expensive labor and time costs. Even sometimes we cannot access label signals due to data privacy and security issues. Thus, exploring high-quality dynamic graph representation learning techniques in an unsupervised manner becomes an important and challenging task.

\begin{figure}
  \centering
  \includegraphics[width=0.9\linewidth]{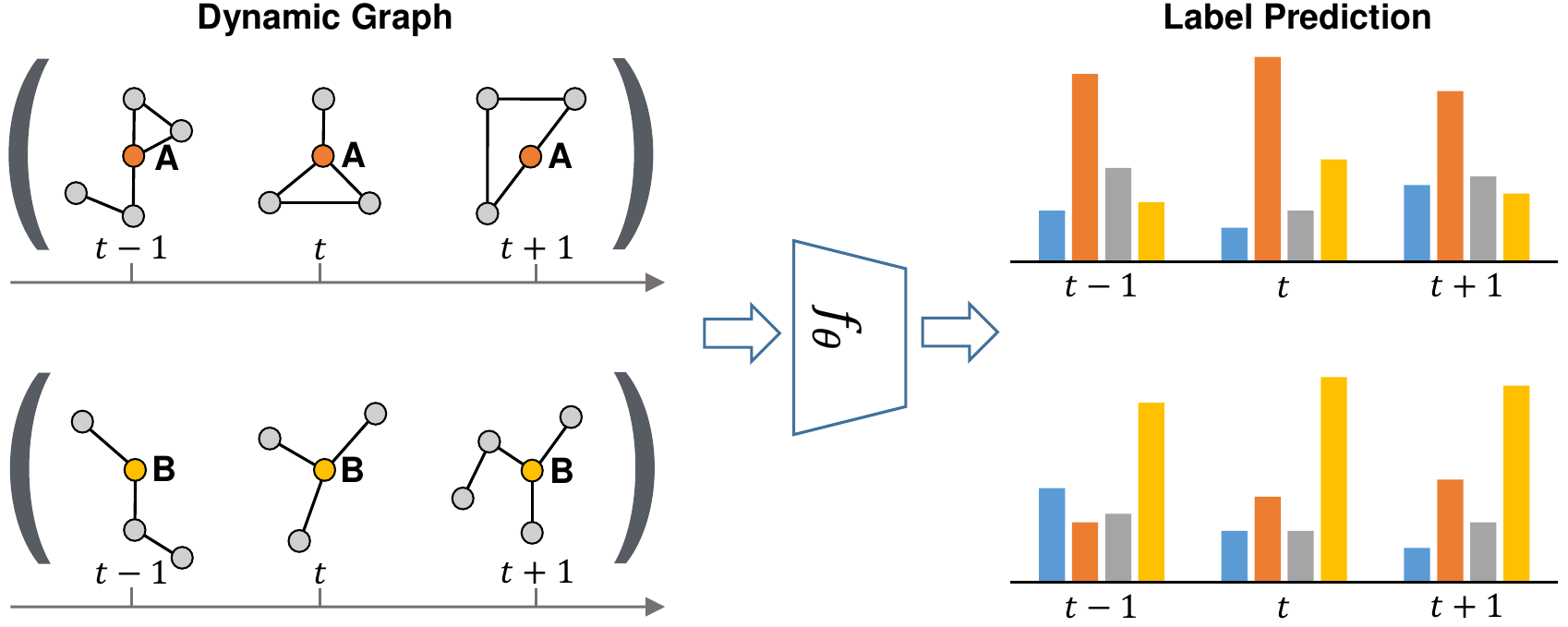}
  \caption{Illustration of our basic idea. In the collected datasets (details are demonstrated in Section~\ref{dataset}), an interesting observation is that the semantics and labels learned by the same node tend to be similar no matter what encoder is used at different timespans of the dynamic graph. We refer to this observation as temporal translation invariance. Based on temporal translation invariance, we claim that the features of node A, on different timespan should be similar and pull apart the features of other nodes, such as B. }
  \label{fig:TTI}
\end{figure}

By conducting empirical analysis on many collected dynamic graphs, as shown in Figure~\ref{fig:TTI}, we find an interesting observation that the prediction labels of the same node tend to keep similarity in different timespans regardless of the encoder employed like GCN~\citep{kipf2016semi}, GAT~\citep{velivckovic2017graph} or MLP~\citep{mcculloch1943logical}. We refer to this phenomenon on dynamic graphs as temporal translation invariance, which is supported by a wealth of intuitions. For instance, most researchers are deeply immersed and dedicated to specific fields throughout their careers. Similarly, the business scope of a company remains almost unchanged in tax transaction networks. The assumption of temporal translation invariance opens the door for us to explore semantic temporal cues through contrastive learning for dynamic graph representation. Specifically, we employ diverse timespan views to create contrastive pairs. By maximizing the semantic mutual information (MI) between identical nodes in contrast views, we can derive informative node representations without the dependence on labeled data. Treating views of different timespans as contrastive pairs is a more graceful way that does not require manual trial-and-error, cumbersome searches, or expensive domain knowledge for augmentation selection~\citep{xia2022simgrace}.

To take advantage of the temporal translation invariance assumption, in this paper, we introduce a Contrastive Learning framework on Dynamic Graphs (CLDG) that is conceptually straightforward and highly effective. To begin with, we employ a timespan view sampling strategy to generate multiple views from continuous dynamic graphs. Then the encoder and projection head are utilized to learn node representations, which is accomplished by effectively aggregating neighborhood information from the adjacency matrix in the timespan view. CLDG treats the semantics of the same node in different timespan views as positive pairs to encourage the representation of the same node in each timespan view to be similar. Note that the adjacency matrix only captures local information from direct neighborhoods and lacks the exploration of implicit topological correlations, which would limit the performance of downstream tasks~\citep{gasteiger2019diffusion,hassani2020contrastive}. For example, in anomaly detection scenarios, fraudsters often deliberately conceal their social relationships through camouflage to reduce their suspicion and evade detection~\citep{dou2020enhancing,ma2021comprehensive}. 
Therefore, to maximize the preservation and extraction of global information in graphs to promote high performance in downstream tasks, we introduce the extended CLDG++ in this work. By incorporating graph diffusion~\citep{page_1999_stanford,kondor_2002_icml} to simulate the propagation process of information on the graph, CLDG++ enlarges the receptive field and effectively models the implicit and global topology of each sampled timespan view. Moreover, we specifically designed a multi-scale contrastive learning objective composed of local-local, local-global, and global-global contrasts for CLDG++ to enhance its simultaneous capture of local and global information. 
Note that rapid or sudden changes in node features or structures on dynamic graphs mostly imply deviations in data patterns and are closely related to anomalies. For example, abnormal companies may suddenly trade in commodities that deviate significantly from their established lines of business or experience a sudden surge in trading activity over a short period. Based on the intuition that anomalies usually violate the assumption of temporal translation invariance, we create an anomaly indicator by capturing contextual consistency and stability in temporal components, which empowers CLDG and its extension to efficiently spot anomalies on dynamic graphs. 
The main contributions are summarized as follows: \looseness=-1

$\bullet$ \textbf{Assumption}: We propose a novel inductive bias named temporal translation invariance on dynamic graphs. Based on this assumption, we generalize the idea of contrastive learning to dynamic graph representation learning by treating nodes in different timespans as contrastive pairs. 

$\bullet$ \textbf{Algorithms}: We propose an efficient contrastive learning method on dynamic graphs to perform representation learning in unsupervised scenarios. In addition to CLDG based on local topological views, we also present its extended version CLDG++ which combines local and global views to derive more comprehensive representations. More interestingly, we seamlessly integrate CLDG/CLDG++ with the anomaly detection task by shaping an anomaly indicator to catch violations of temporal translation invariance.

$\bullet$ \textbf{Downstream Applications}: We corroborate the effectiveness of CLDG and CLDG++ on node classification and dynamic graph anomaly detection tasks. The experiments showcase the remarkable potential of our methods in leveraging large-scale unlabeled dynamic graph data.

Our preliminary work is published in~\citep{xu2023cldg}. Compared to the previous version, we first introduce a novel variant called CLDG++, which leverages Personalized PageRank (PPR)~\citep{page_1999_stanford} and heat kernel~\citep{kondor_2002_icml} to uncover the global contextual relationships and the degree of interaction between nodes in the graph. This approach goes beyond the limited local receptive field characterized by simple binary connections in the adjacency matrix, allowing us to capture more comprehensive insights into implicit links and global topological properties. For example, recovering social relations maliciously concealed by fraudsters facilitates exposing underlying anomalous behaviors in anomaly detection scenarios. To effectively capture local and global information in large-scale unlabeled data, we design a multi-scale contrastive learning objective composed of local-local, local-global, and global-global contrasts accordingly. To promote the application research of the proposed representation model, we create an anomaly indicator to catch violations of temporal translation invariance based on the intuition that rapid or sudden changes in node features or structures on dynamic graphs are closely related to anomalies. Then, CLDG and CLDG++ are successfully applied to unsupervised dynamic graph anomaly detection, which has broad applications in many high-impact domains but has received relatively little scrutiny.

\section{Related Work}
\label{sec:related_work}
\subsection{Contrastive Learning on Graphs}
Supervised and semi-supervised learning still dominate in graph neural networks~\citep{feng2020graph, wang2020graph, dornaika2023unified, han2023dual}. However, the graph itself is an abstraction of the real world, and graph data suffers from the problems of lack of labels and difficulty in labeling. Recently, contrastive learning is highly successful in computer vision (CV) and natural language processing (NLP)~\citep{wu2018unsupervised, van2018representation, he2020momentum, chen2020simple, grill2020bootstrap, gao2021simcse}. Inspired by the above methods, there are some works that extend contrastive learning to graphs. DGI~\citep{velickovic2019deep} extends deep InfoMax~\citep{hjelm2018learning} to graphs and maximizes MI between global graph embeddings and local node embeddings. GraphCL~\citep{you2020graph} proposes a novel graph contrastive learning framework that systematically explores the performance impact of various combinations of four different data augmentations. GCA~\citep{zhu2021graph} and GRACE~\citep{zhu2020deep} pay more attention to the graph data augmentation and propose adaptive data augmentation schemes on topology and node attributes. MVGRL~\citep{hassani2020contrastive} creates another view for the graph by introducing graph diffusion~\citep{gasteiger2019diffusion}. A discriminator contrasts node representations from one view with graph representation of another view and vice versa, and scores the agreement between representations, which is used as the training signal. CCA-SSG~\citep{zhang2021canonical} first generates two views of the input graph through data augmentation, and then uses the idea based on Canonical Correlation Analysis (CCA) to maximize the correlation between the two views and encourages different feature dimensions to capture distinct semantics. However, the above methods have two problems: (1) They require two views generated by corrupting the original graph, such as 
node dropping, edge perturbation, and attribute masking, etc. Inappropriate data augmentation may introduce noisy information resulting in semantic and label changes. (2) They are designed for static graphs. Applying contrastive learning directly to dynamic graphs is not straightforward. 

\subsection{Representation Learning on Dynamic Graphs}
Traditionally, the static graph representation problem is intensively studied by researchers and a variety of effective works are proposed~\citep{kipf2016semi, velivckovic2017graph, chien2021node,zhao2023learnable,jiang2023dropagg,xu2025text}. However, real-world networks all evolve over time, which poses important challenges for learning and inference. Therefore, there has been an increasing amount of research on dynamic graph representations recently~\citep{ma2023dygl}. According to the modeling method of the dynamic graph, these works can be roughly divided into discrete-time methods~\citep{goyal2018dyngem, sankar2020dysat, xu2020inductive, pareja2020evolvegcn, xue2020modeling, wang2021temporal} and continuous-time methods~\citep{zuo2018embedding, kumar2019predicting, trivedi2019dyrep, lu2019temporal}.

\noindent\textbf{Discrete-time Methods} 
DynGEM~\citep{goyal2018dyngem} uses a deep autoencoder to capture the connectivity trends in a graph snapshot at any time step. DySAT~\citep{sankar2020dysat} generates dynamic node representations through self-attention along both structural and temporal. TGAT~\citep{xu2020inductive} uses the self-attention mechanism and proposes a time encoding technique based on the theorem of Bochner. STAR~\citep{xu2019spatio}, DyHATR~\citep{xue2020modeling}, dyngraph2vec~\citep{goyal2020dyngraph2vec} and TemGNN~\citep{wang2021temporal} use different encoders (such as GCN~\citep{kipf2016semi}, autoencoder~\citep{baldi2012autoencoders}, hierarchical attention model, etc.) to extract the features of each time snapshot respectively, and then introduce sequence models such as LSTM~\citep{hochreiter1997long} and GRU~\citep{cho2014learning} to capture timing information. 

\noindent\textbf{Continuous-time Methods} 
HTNE~\citep{zuo2018embedding} proposes a Hawkes process based temporal network embedding method to capture the influence of historical neighbors on the current neighbors. JODIE~\citep{kumar2019predicting} uses a coupled RNN architecture to update the embedding of users and items at every interaction. DyRep~\citep{trivedi2019dyrep} builds a two-time scale deep temporal point process approach to capture the continuous-time fine-grained temporal dynamics processes. M$^2$DNE~\citep{lu2019temporal} designs a temporal attention point process to capture the fine-grained structural and temporal properties in micro-dynamics, and defines a dynamics equation to impose constraints on the network embedding in macro-dynamics.

However, existing dynamic graph work still suffers from at least one of the following limitations: (1) 
Most sequence models such as RNN-like methods have high time costs and space complexity, and are not easily parallelized. This hinders the scaling of existing dynamic graph models to large-scale graphs. (2) Most models by reconstructing future states or temporal point processes or sequences may learn noisy information as they try to fit each new interaction in turn. (3) The temporal regularizer is similar to contrastive learning without negative examples, which forces the node representation to smooth from adjacent snapshots. However, a potential problem of this approach is the existence of completely collapsed solutions.

\subsection{Anomaly Detection on Graphs}
In recent years, the widespread application of graphs in domains such as finance, social networks, security, and medicine has attracted considerable interest in research on graph anomaly detection~\citep{zheng2023survey,shi2023edge}. Graph anomaly detection aims to identify abnormal objects (i.e., nodes, edges, and sub-graphs) that deviate from the majority in graph data, which plays a crucial role in several high-impact applications, including financial fraud detection, fake account detection, and network intrusion detection~\citep{ma2021comprehensive,zhang2024graph,xu2025revisiting}. At present, most existing works focus on node-level detection. For instance, 
% DOMINANT
DOMINANT~\citep{ding2019deep} detects anomalies by measuring the reconstruction errors of nodes from both the structural and attribute perspectives after GCN.
% AEGIS
AEGIS~\citep{ding2021inductive} first learns anomaly-aware node representations through an autoencoder network, then employs generative adversarial learning to detect anomalies among data.
% CoLA
CoLA~\citep{liu2021anomaly} first introduces contrastive learning to effectively detect node anomalies in the graph, by measuring whether the target node matches its positive and negative neighbor sample pairs.
% ANEMONE
Based on CoLA, ANEMONE~\citep{jin2021anemone} adds patch-level (i.e., node versus node) consistency to further estimate the anomaly score of nodes.
% GRADATE
The subgraph-subgraph contrast first proposed by GRADATE~\citep{duan2023graph} is combined with the widely used node-subgraph and node-node contrasts to capture anomaly information.
However, the above methods all work under the settings of static graphs. In contrast, anomaly detection on dynamic graphs has received relatively little scrutiny. Only a few works have explored dynamic graph scenarios with temporal information.
% NetWalk
NetWalk~\citep{yu2018netwalk} utilizes random walks to generate node embeddings and then flag anomalies based on a dynamic clustering model.
% TADDY
TADDY~\citep{liu2021anomalytkde} detects edge anomalies through a dynamic graph transformer model.
% SAD
SAD~\citep{tian2023sad} proposes a semi-supervised dynamic graph anomaly detection framework, which also uses a time-equipped memory bank and pseudo-label contrastive learning to exploit the potential of unlabeled samples.

In summary, the existing research on anomaly detection in dynamic graphs is limited. Our study is committed to combining unsupervised dynamic graph representation methods with graph anomaly detection tasks to provide new ideas for successfully identifying dynamic graph anomalies.

\newcommand{\tabincell}[2]{\begin{tabular}{@{}#1@{}}#2\end{tabular}} 
\begin{table}\small
\centering
\renewcommand\arraystretch{1.2} 
\setlength\tabcolsep{20pt}
  \caption{Notations and Descriptions.}
  \label{tab:ND}
  \begin{tabular}{cc}
    \toprule
    Notations & Descriptions\\
    \midrule
    $\mathcal{G}$ & A dynamic graph \\
    $\mathcal{V}$ & The set of nodes in $\mathcal{G}$ \\
    $\mathcal{E}$ & The set of edges in $\mathcal{G}$ \\
    $\tau$  & The temperature parameter \\
    $s$ & The view timespan factor \\
    $v$ & The number of views sampled \\
    $a,\mathbf{a},\mathbf{A}$ & Scalar, vector, matrix\\
    $\mathbb{R}_{\textrm{sample}}\left ( \cdot ,\cdot ,\cdot  \right )$ & Timespan view sampling function \\
  \bottomrule
\end{tabular}
\end{table}

\section{Preliminaries}
This section presents the necessary notations, definitions, and studied problems in this paper. The main notations are summarized in Table~\ref{tab:ND}.

\subsection{Dynamic Graph Modeling}
The existing dynamic graph modeling methods can be roughly divided into two categories~\citep{xue2022dynamic, kazemi2020representation}: discrete-time dynamic graph and continuous-time dynamic graph. We formally define two modeling methods as follows:

\begin{definition} [\textbf{Discrete-time Dynamic Graph}]
A discrete-time dynamic graph (DTDG) is a sequence of network snapshots within a given time interval. Formally, we define a DTDG as a set $\left \{\mathcal{G}^{1}, \mathcal{G }^{2},... ,\mathcal{G }^{T} \right \}$ where $\mathcal{G}^{t}=\left \{\mathcal{V}^{t}, \mathcal{E}^{t}\right \}$ is the graph at snapshot $t$, $\mathcal{V}^t$ is the set of nodes in $\mathcal{G}^{t}$, and $\mathcal{E}^t\subseteq \mathcal{V}^t\times \mathcal{V}^t$ is the set of edges in $\mathcal{G}^{t}$.
\end{definition}

\begin{definition} [\textbf{Continuous-time Dynamic Graph}]
A continuous-time dynamic graph (CTDG) is a network with edges and nodes annotated with timestamps. Formally, we define a CTDG as $\mathcal{G}=\left \{\mathcal{V}^T, \mathcal{E}^T, \mathcal{T}\right \}$ where $\mathcal{T}:\mathcal{V},\mathcal{E}\rightarrow \mathbb{R}^{+}$ is a function that maps each node and edge to a corresponding timestamp.
\end{definition}

\subsection{Problem Formulation}
The objective of this paper is to design a dynamic graph representation method. The mathematical formulation of this problem could be defined as:
\begin{problem} [\textbf{Representation Learning on Dynamic Graph}]
Given a dynamic graph $\mathcal{G}=\left \{\mathcal{V}, \mathcal{E}, \mathcal{T}\right \}$, our goal is to learn a mapping function $f:v_{i}\to \mathbf{z}_{i}\in \mathbb{R}^{d}$, where $d\ll \left | \mathcal{V}\right |$, and $\mathbf{z}_{i}$ is the embedded vector that preserves both temporal and structural information of vertex $v_{i}$.
\end{problem} 

With the aforementioned notations, we formalize the problem of dynamic graph anomaly detection as follows:
\begin{problem} [\textbf{Anomaly Detection on Dynamic Graph}]
Given a dynamic graph $\mathcal{G}$, the goal of anomaly detection is to learn an anomaly score function $f$ to calculate the anomaly score $s_{i}=f\left ( v_{i} \right )$ of each node. The anomaly score $s_{i}$ measures the degree of abnormality for node $v_i$. A larger anomaly score means that it is more likely to be anomalous.
\end{problem} 

\section{Proposed Method}
\begin{figure*}
  \centering
  \includegraphics[width=1\linewidth]{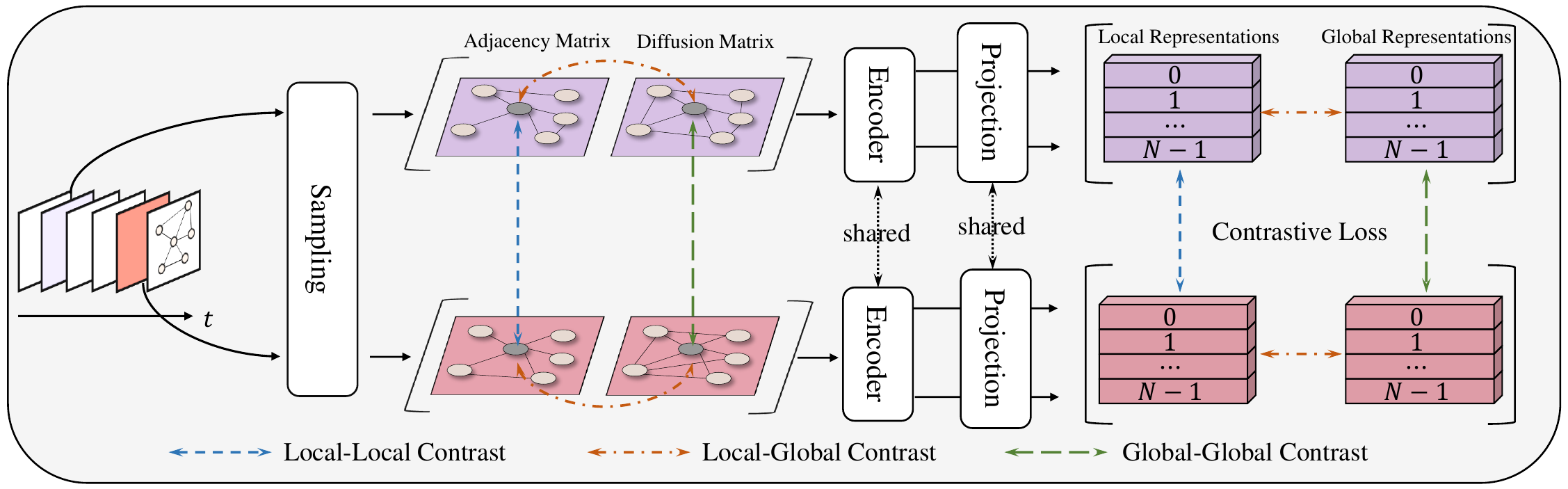}
  \caption{The architecture of the CLDG++. Given an input graph, we first sample multiple views through a timespan view sampling layer. Subsequently, each view generates a diffusion matrix through graph diffusion, where the adjacency matrix and diffusion matrix offer local and global perspectives of the graph structure, respectively. Then, the adjacency and diffusion matrices of the views are fed into the encoder to generate node embeddings, and a projection head is utilized to map the embeddings into the space of the contrastive loss. Finally, CLDG trains the model by maintaining the temporal translation invariance between local-local contrast, while CLDG++ adds both local-global contrast and global-global contrast to capture both local and global topological properties.}
  \label{fig:overview}
\end{figure*}

In this section, we propose a new unsupervised dynamic graph representation learning method CLDG++, whose framework overview and workflow are shown in Figure~\ref{fig:overview}. It consists of five major lightweight components: timespan view sampling layer, graph diffusion layer, base encoder, projection head and contrastive loss function. To begin with, the view sampling layer extracts the temporally-persistent signals. 
Then, the global view is captured through the graph diffusion layer. Furthermore, the base encoder learns the local and global representations, and the projection head maps the representations into the space of the contrastive loss. Finally, the contrastive loss function is used to maintain the temporal translation invariance of the local-local, local-global, and global-global.

\begin{figure*}
	\centering
    \subfloat[DBLP]{\label{fig:emp1}\includegraphics[width=0.333\linewidth]{
    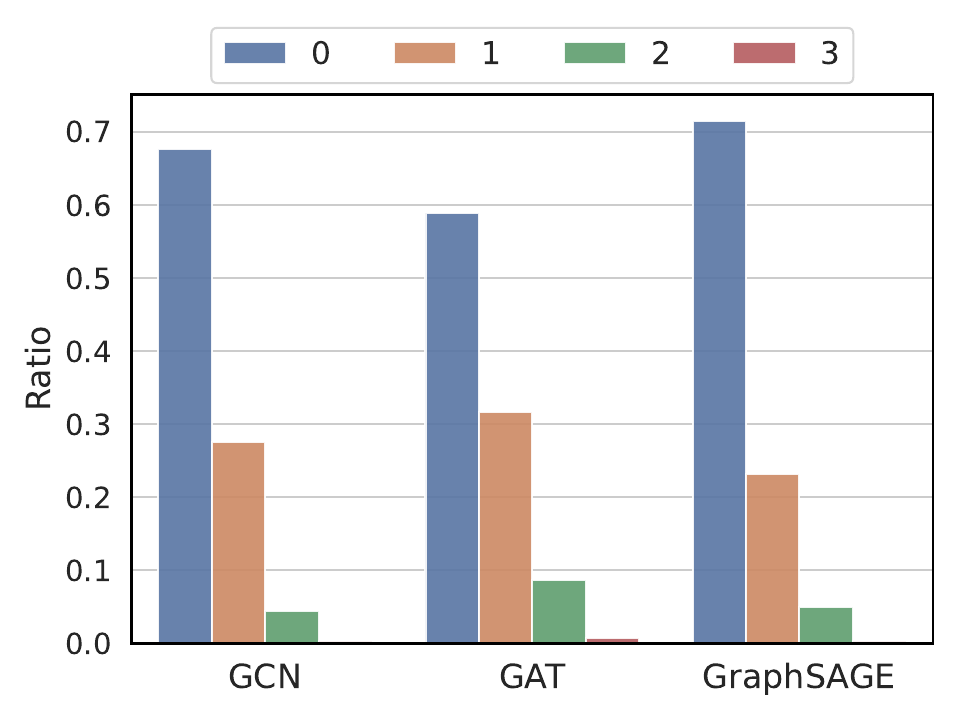}}
    \subfloat[TAX]{\label{fig:emp2}\includegraphics[width=0.333\linewidth]{
    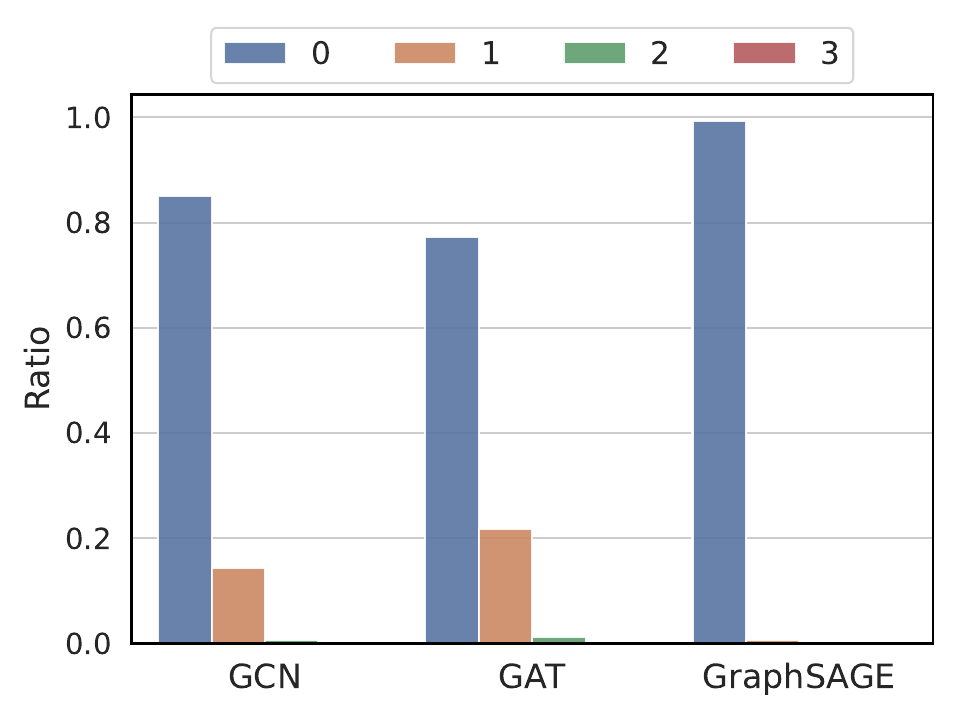}}
    \subfloat[BITalpha]{\label{fig:emp3}\includegraphics[width=0.333\linewidth]{
    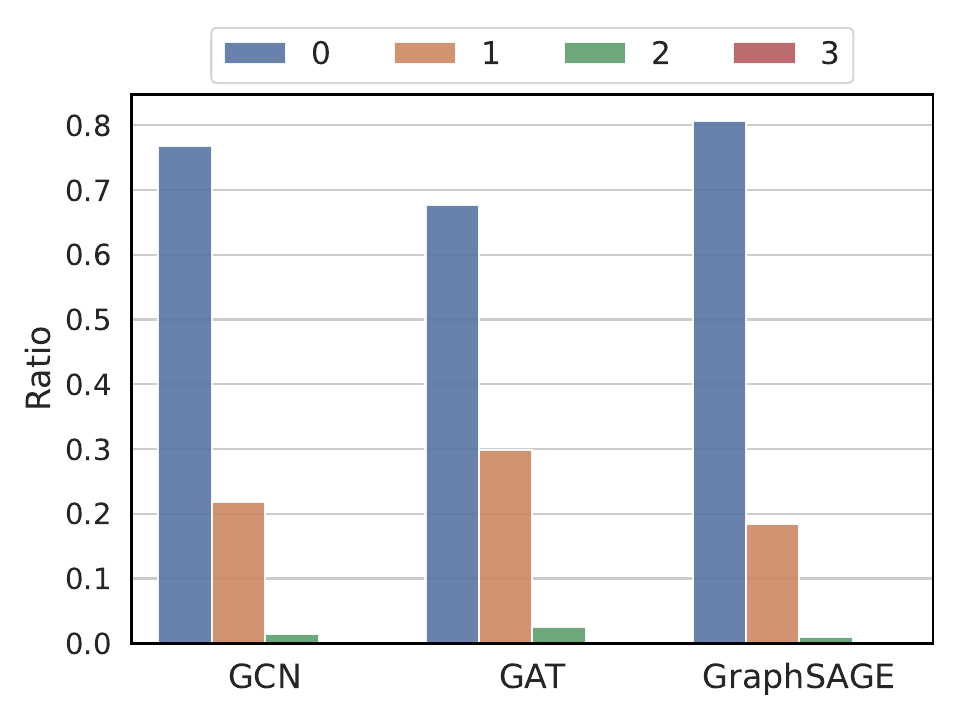}}
    \caption{The ratio of the number of times the predicted node label has changed across timespans.}
    \label{fig:empirical}    
\end{figure*}

\subsection{Temporal Translation Invariance}\label{empirical}
To explore the properties of dynamic graphs, we perform some empirical studies. Specifically, we first process dynamic graph datasets $\mathcal{G}$ by spliting each of them into many timespans $\left \{\mathcal{G}^{1}, \mathcal{G }^{2},... ,\mathcal{G }^{T} \right \}$ (the specific datasets described in Section~\ref{dataset}), where we set $T$ to 4. For each timespans, we train common GNN encoders with the same architecture but without shared weights, where we choose three GNN encoders: GCN, GAT and GraphSAGE. The trained model is then used to predict the node labels. We calculate the ratio of the number of times the predicted node label changed over the timespan on three representative datasets (referred to as DBLP, TAX, and BITalpha), as illustrated in Figure~\ref{fig:empirical}. For example, a change count of 0 indicates consistent label predictions across all four timespans, and the Y-axis indicates the ratio of nodes demonstrating this behavior relative to the total number of nodes. An interesting observation is that regardless of the encoder used for training, the prediction labels of the same node tend to be similar in different timespans. We refer to this phenomenon as temporal translation invariance. 

Under the assumption of temporal translation invariance, an unsupervised opportunity and pathway are provided to more elegantly exploit the supplementary temporal cues inherent in dynamic graphs. We utilize different timespan views to construct contrastive pairs. Specifically, we treat the semantics of the same node in different timespan views as positive pairs, pulling closer the representation of the same node in each timespan view, and pulling apart the representation of different nodes. This method effectively utilizes temporal information while avoiding noisy information caused by graph data augmentation techniques, such as node dropping, edge perturbation, and attribute masking commonly used in existing graph contrastive learning methods.

\begin{figure}
  \centering
  \includegraphics[width=0.8\linewidth]{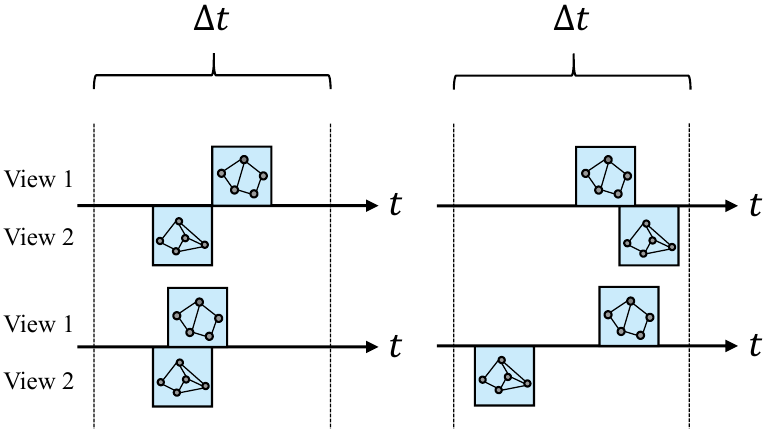}
  \caption{Four candidate timespan view sampling strategies.}
  \label{fig:sampling}
\end{figure}

\subsection{Timespan View Sampling Layer}\label{tpgs}
Based on temporal translation invariance, we utilize different timespan views as contrastive pairs. However, how to specifically choose appropriate timespans is still an open question. First, it is intuitive that the interval distance between different timespan views may affect the contrastive learning results. In the case of two timespan views, if more temporal overlap exists between two views, they theoretically share more similar semantic contexts, but this may lead to over-simplified tasks. If two views are separated for a long time, they may have completely different neighbors and may no longer share the same semantic context. In addition, how to choose the appropriate number and size of timespan views is also to be studied. Therefore, we design four different timespan view sampling strategies to explore the optimal view interval distance selection, as shown in Figure~\ref{fig:sampling}. The main difference between each strategy is the rate of physical temporal overlap, thereby sharing different semantic contexts. Meanwhile, each sampling strategy considers both the number and size of views. Next, we formally define the timespan view sampling layer.

Specifically, given a continuous-time dynamic graph $\mathcal{G}$, we first define that the overall timespan of graph $\mathcal{G}$ is $\Delta t=\max{\left ( \mathcal{T} \right )}-\min{\left ( \mathcal{T} \right )}$. Then, we set two factors $v,s$ to control the number and size of the sampling timespan views respectively, i.e., $v$ views are sampled for each strategy and the timespan of each view is $\Delta t / s$, and $v,s\in \mathbb{N^{*}}$ are hyperparameters that can be set according to the actual physical implication of the graph and the density of the graph. Finally, the timespan view sampling strategy can be formulated as:
\begin{equation}
\left ( \mathcal{T}_{1},\cdots ,\mathcal{T}_{v} \right )=\mathbb{R}_{\textrm{sample} }\left (s ,v,\Delta t \right ),
\end{equation}
where $\mathcal{T}_{i}\in \left [ \min{\left ( \mathcal{T} \right )}+\frac{\Delta t}{2s}, \max{\left ( \mathcal{T} \right )}-\frac{\Delta t}{2s} \right ], \forall i\in \left [ 1,v \right ]$, and $\mathbb{R}_{\textrm{sample} }\left ( \cdot ,\cdot ,\cdot  \right )$ returns a sample time tuple.

We obtain $v$ timespan views, i.e., $\widetilde{\mathcal{G}}_{1}, \widetilde{\mathcal{G}}_{2},\cdots, \widetilde{\mathcal{G}}_{v}$, according to the sample time tuples $\left ( \mathcal{T}_{1},\cdots ,\mathcal{T}_{v} \right )$, where $\widetilde{\mathcal{G}}_{i}$ retains all edges of $\mathcal{G}$ in  a specific timeframe $\left [ \mathcal{T}_{i}-\frac{\Delta t}{2s},\mathcal{T}_{i}+\frac{\Delta t}{2s} \right ]$.
Specifically, the sample time tuple can be generated by four different strategies. 
We formally define four sampling strategies by the time difference between $\mathcal{T}_{i}$ in the sample time tuple and its predecessor $\mathcal{T}_{i-1}$ and successor $\mathcal{T}_{i+1}$. 

\textbf{Sequential Sampling Strategy}. 
This strategy first divides the dynamic graph into $s$ timespan views, with no intersection between each view, and then randomly samples $v$ unduplicated views, where $v\leq s$. This strategy is similar to the processing method of DTDG, which can be directly used in DTDG. 
\begin{equation}
\left| \mathcal{T}_{i}-\mathcal{T}_{i\pm 1} \right| = \alpha \frac{\Delta t}{s},
\end{equation}
where $\alpha \in \mathbb{N^{*}}$, the time difference between any two views is an integer multiple of the timespan of the view. 

\textbf{High Overlap Rate Sampling Strategy}.
The $v$ views sampled by this strategy are interdependent. We set an overlap rate of 75\%, that is, the time between the sampled dynamic graphs $\widetilde{\mathcal{G}}_{i}$ and $\widetilde{\mathcal{G}}_{i\pm1}$ corresponding to $\mathcal{T}_{i}$ and $\mathcal{T}_{i\pm 1}$ has a 75\% overlap. 
\begin{equation}
\left| \mathcal{T}_{i}-\mathcal{T}_{i\pm 1} \right| = \frac{\Delta t}{4s},
\label{eq:high_overlap}
\end{equation}
where $\mathcal{T}_{1}$ is limited by $\left [ \min{\left ( \mathcal{T} \right )}+\frac{\Delta t}{2s},\max{\left ( \mathcal{T} \right )}-\frac{\left ( 2+v \right )\cdot \Delta t}{4s} \right ]$ and random sampling in each epoch, assuming that $\mathcal{T}_{1}< \mathcal{T}_{i} < \mathcal{T}_{v}$ is satisfied in the sampled time tuple. Then $\mathcal{T}_{2}$ to $\mathcal{T}_{v}$ are obtained according to Eq.~\ref{eq:high_overlap}.

\textbf{Low Overlap Rate Sampling Strategy}.
The $v$ views sampled by this strategy are interdependent. We set an overlap rate of 25\%, which means that the time between the sampled dynamic graphs $\widetilde{\mathcal{G}}_{i}$ and $\widetilde{\mathcal{G}}_{i\pm1}$ corresponding to $\mathcal{T}_{i}$ and $\mathcal{T}_{i\pm 1}$ has a 25\% overlap. 
\begin{equation}
\left| \mathcal{T}_{i}-\mathcal{T}_{i\pm 1} \right| = \frac{3\Delta t}{4s},
\label{eq:low_overlap}
\end{equation}
where $\mathcal{T}_{1}\in \left [ \min{\left ( \mathcal{T} \right )}+\frac{\Delta t}{2s},\max{\left ( \mathcal{T} \right )}-\frac{\left ( 2+3v \right )\cdot \Delta t}{4s} \right ]$ and random sampling in each epoch, assuming that $\mathcal{T}_{1}< \mathcal{T}_{i} < \mathcal{T}_{v}$ is satisfied in the sampled time tuple. Then $\mathcal{T}_{2}$ to $\mathcal{T}_{v}$ are obtained according to Eq.~\ref{eq:low_overlap}.

\textbf{Random Sampling Strategy}.
$\mathbb{R}_{\textrm{sample} }\left ( \cdot ,\cdot ,\cdot  \right )$ randomly returns a set of sample time tuples, where $\forall i $ satisfies $\mathcal{T}_{i}\in \left [ \min{\left ( \mathcal{T} \right )}+\frac{\Delta t}{2s},\max{\left ( \mathcal{T} \right )}-\frac{\Delta t}{2s} \right ]$. The dynamic graphs $\widetilde{\mathcal{G}}_{i}$ and $\widetilde{\mathcal{G}}_{j}$ sampled corresponding to $\mathcal{T}_{i}$ and $\mathcal{T}_{j}$ may not overlap or partially overlap in time.
\begin{equation}
\left| \mathcal{T}_{i}-\mathcal{T}_{i\pm 1} \right| \in \left [ 0,\Delta t - \frac{\Delta t}{s} \right ].
\end{equation}

Finally, the timespan view sampling layer samples $V$ graph views in each training epoch, defined as $\left\{\widetilde{\mathcal{G}}_{1},\widetilde{\mathcal{G}}_{2},\cdots ,\widetilde{\mathcal{G}}_{V} \right\}$ and $\widetilde{\mathcal{G}}_{i}=\left\{\mathbf{X}_{i},\mathbf{A}_{i}\right\}$, where $\mathbf{X}_{i}$ and $\mathbf{A}_{i}$ are the feature matrix and adjacency matrix of the i-th view. 

\begin{figure*}[!ht]
\centering
    \subfloat[]{\label{fig:heat1}\includegraphics[width=0.4\linewidth]{
    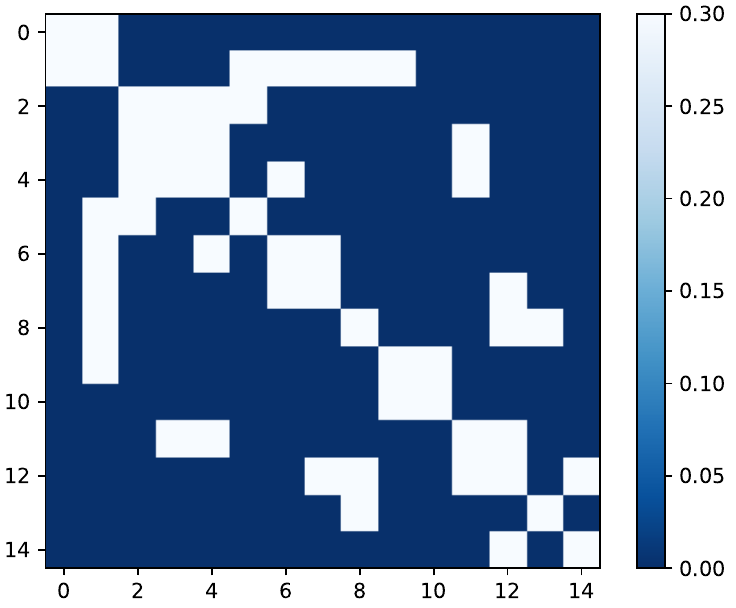}}
    \subfloat[]{\label{fig:heat2}\includegraphics[width=0.4\linewidth]{
    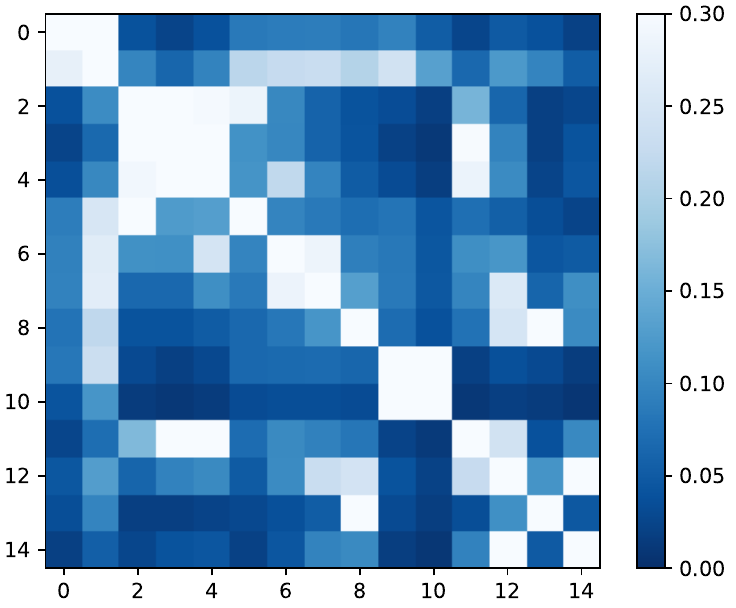}}\\
\centering
    \subfloat[]{\label{fig:heat3}\includegraphics[width=0.4\linewidth]{
    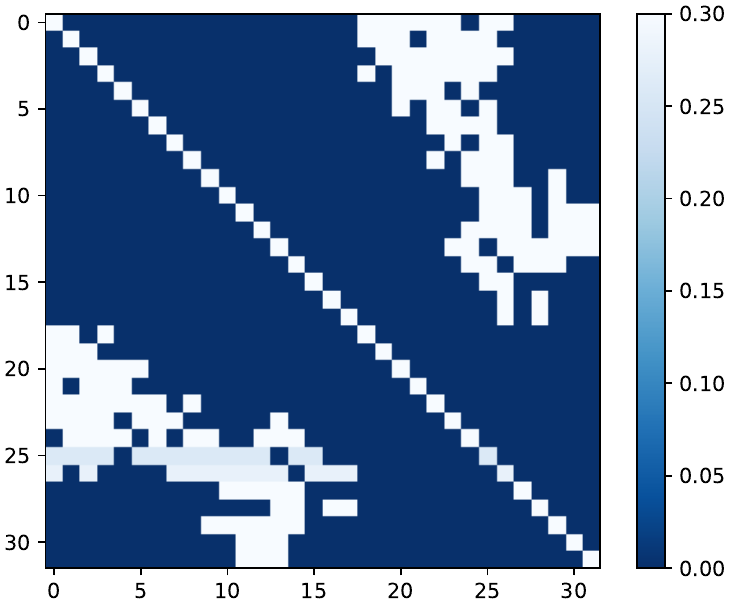}}
    \subfloat[]{\label{fig:heat4}\includegraphics[width=0.4\linewidth]{
    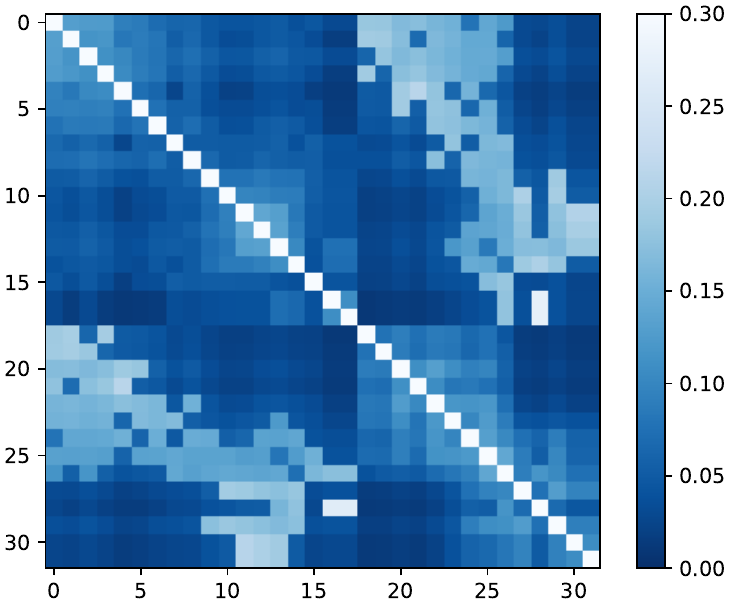}}
    \caption{The heatmaps (a) and (c) of the normalized adjacency matrix, and the heatmaps (b) and (d) of the normalized diffusion matrix are visualized for the Florentine families graph~\citep{breiger1986cumulated} and the Davis Southern women social network~\citep{davis1941deep}, respectively.}
    \label{fig:heat}
\end{figure*}

\subsection{Graph Diffusion Layer}
Based on the above sampling strategy, we transform the original dynamic graph into timespan views, which define the topology of each view by the adjacency matrix. However, adjacency matrices only describe local information about simple binary connectivity from direct neighborhoods~\citep{gasteiger2019diffusion,hassani2020contrastive}, and lack the consideration of implicit topological correlations, which would limit the performance of downstream tasks. In contrast, preserving and extracting as much global information as possible from graphs facilitates high-quality representation learning. We introduce two popular instantiations of generalized graph diffusion, PPR~\citep{page_1999_stanford} and heat kernel~\citep{kondor_2002_icml} to uncover the global contextual relationships and the degree of interaction between nodes in the graph by simulating the propagation process of information on the graph. 
As shown in Figure~\ref{fig:heat}, we visualize the adjacency matrices of the Florentine families graph~\citep{breiger1986cumulated} and the Davis Southern women social network~\citep{davis1941deep}, and the diffusion matrix generated by PPR, respectively. It is obvious from Figure~\ref{fig:heat2} and Figure~\ref{fig:heat4} that the diffusion matrix depicts more potential relationships between nodes, eliminating the restriction of using only direct neighbors. By aggregating information from larger neighbors within the diffusion matrix, the model is enabled to learn global semantic information. Mathematically, the closed-form solution of the PPR and heat kernel is defined as:

\begin{equation}
    \label{eq:ppr}
\begin{aligned}
	\mathbf{S}^{\textrm{PPR}}=\alpha\left(\mathbf{I}_n-(1-\alpha) \mathbf{D}^{-1/2}\mathbf{A}\mathbf{D}^{-1/2}\right)^{-1},
    \end{aligned}
\end{equation}

\begin{equation}
    \label{eq:heat}
	\mathbf{S}^{\textrm{heat}}= \exp{\left({t\mathbf{A}\mathbf{D}^{-1}-t}\right)},
\end{equation}
where $\mathbf{D}$ is the degree matrix of $\mathbf{A}$. $\alpha$ denotes teleport probability in a random walk and $t$ is diffusion time. For the $V$ local timespan views sampled, we generate corresponding global diffusion views $\left\{\mathbf{S}_{1},\mathbf{S}_{2},\cdots ,\mathbf{S}_{V} \right\}$.

\subsection{Base Encoder Layer}
For local and global timespan views, to ensure that the learned representations have better consistency between each class of view, we utilize encoder architecture without shared weights to extract local and global information.  Specifically, for simplicity, we choose GCN~\citep{kipf2016semi} as the basic encoder to model structural dependencies, and we feed all local semantic timespan views into the encoder $f_{l}$ and all global semantic timespan views into encoder $f_{g}$. The node representations are updated by multi-layer graph convolution: 
\begin{align}
\label{eq:gcn1}
\mathbf{H}^{l} &= f_{l}\left ( \mathbf{A},\mathbf{X} \right ), \\
\label{eq:gcn2}
\mathbf{H}^{g} &= f_{g}\left ( \mathbf{S},\mathbf{X} \right ), 
\end{align}
where $\mathbf{H}^{l} \in \mathbb{R}^{N \times d}$ and $\mathbf{H}^{g} \in \mathbb{R}^{N \times d}$ are the node representations learned from $ \mathbf{A}$ and $ \mathbf{S}$, respectively.
It is worth noting that the different views sampled on the dynamic graph according to the timespan view sampling layer contain different sets of nodes and edges. Therefore, we used the minibatch training approach in the specific implementation. 

\subsection{Projection Head Layer}
The base encoder extracts the representation of nodes from the local and global timespan views. Then, we feed the representations containing local and global information respectively into a projection head, which maps the representation to the space where contrastive loss is applied. The projection head formula is as follows: 
\begin{equation}
\mathbf{z}_{i,l}^{\mathcal{T}},\mathbf{z}_{i,g}^{\mathcal{T}}=\textrm{proj} \left (  \mathbf{h}_{i,l}^{\mathcal{T}} \right ),\textrm{proj} \left (  \mathbf{h}_{i,g}^{\mathcal{T}} \right ) ,
\end{equation}
where $\mathbf{h}_{i,l}^{\mathcal{T}}$ and $\mathbf{h}_{i,g}^{\mathcal{T}}$ are the local and global representations of node $i$ extracted under the timespan view $\mathcal{T}$, respectively. $\textrm{proj} \left ( \cdot  \right ) $ is composed of a fully connected layer, $l_{2}$ normalized and LeakyReLU activation function.

\subsection{Contrastive Loss Function}
Based on the observation that node labels tend to be invariant over the entire dynamic graph time period, we propose a new inductive bias on dynamic graphs, i.e., temporal translation invariance. Distinguishing from previous work, we learn node embedding by maximizing the temporal consistency between local-local, local-global, and global-global semantic views.
Specifically, in temporal translation invariance, we treat the semantics between locals or globals of the same/different node in different timespan views and the semantics between local and global of the same/different node in the same timespan view as positive/negative pairs.
CLDG++ is trained to minimize temporal translation invariance. The contrastive learning loss of CLDG++ is defined as follows: 

\begin{equation}
\begin{split}
\mathcal{L} = &\sum_{i=1}^{N}\sum_{q=1}^{V}\sum_{k\neq q}^{V} \Bigg( \textrm{ctr}\left( \mathbf{z}_{i,l}^{\mathcal{T}_{q}},\mathbf{z}_{i,l}^{\mathcal{T}_{k}} \right) + \textrm{ctr}\left( \mathbf{z}_{i,g}^{\mathcal{T}_{q}},\mathbf{z}_{i,g}^{\mathcal{T}_{k}} \right) \\
&\quad + \textrm{ctr}\left( \mathbf{z}_{i,l}^{\mathcal{T}_{q}},\mathbf{z}_{i,g}^{\mathcal{T}_{q}} \right) + \textrm{ctr}\left( \mathbf{z}_{i,l}^{\mathcal{T}_{k}},\mathbf{z}_{i,g}^{\mathcal{T}_{k}} \right) \Bigg)
\end{split}
\label{eq:loss}
\end{equation}
where $N$ is the batch size, and $V$ is the multiple views sampled by the timespan view sampling layer. $\textrm{ctr}\left ( \cdot ,\cdot  \right )$ is InfoNCE~\citep{van2018representation} contrastive approach, which is defined in Eq.~\ref{eq:infonce}. CLDG exclusively focuses on local-local contrast, resulting in the loss of node $i$ degenerates to $\textrm{ctr}\left( \mathbf{z}_{i,l}^{\mathcal{T}_{q}},\mathbf{z}_{i,l}^{\mathcal{T}_{k}} \right)$.

\begin{equation}
\textrm{ctr}\left( \mathbf{z}_{i,\cdot}^{\mathcal{T}_{q}},\mathbf{z}_{i,\cdot}^{\mathcal{T}_{k}} \right)=-\log\frac{\exp{\left ( \mathbf{z}_{i,\cdot}^{\mathcal{T}_{q}}\cdot \mathbf{z}_{i,\cdot}^{\mathcal{T}_{k}}/\tau  \right )}}{\sum_{j=1}^{N}\exp{\left ( \mathbf{z}_{i,\cdot}^{\mathcal{T}_{q}}\cdot \mathbf{z}_{j,\cdot}^{\mathcal{T}_{k}}/\tau  \right )}},
\label{eq:infonce}
\end{equation}
where $\tau$ is a temperature parameter.

To sum up, in each training epoch, we first sample multiple different views through the timespan view sampling layer. Subsequently, a global diffusion timespan view is generated by employing PPR and heat kernel on the sampled local timespan views.
Afterward, the local and global node representations in the minibatch are learned by using the encoder and projection head. Finally, the parameters in the encoder and projection head are updated by minimizing Eq.~\ref{eq:loss}. 

\subsection{Anomaly Discriminator}
CLDG and CLDG++ learn good node representation through temporal translation invariance. However, we realize that anomalous nodes may violate the assumption of temporal translation invariance. Instances such as an anomalous company exhibiting a substantial deviation in traded goods from its established business scope over a period of time or experiencing a sudden surge in trading activities over a short period of time. Capturing such anomalies would make it possible to extend CLDG and CLDG++ to anomalous node detection. A new challenge is how to accurately quantify each node-specific anomaly score from the learned low-dimensional node features without a given anomalous node.

To address the above challenges, we design an anomaly discriminator in the inference phase. Specifically, after the model training is completed, the $V$ timespan views are first sampled in the inference phase using the sequential sampling strategy. Then node $i$ is encoded using the base encoder and the projection head, the learned representation of node $i$ as $ \left [\mathbf{z}_{i,l}^{\mathcal{T}_{1}},\mathbf{z}_{i,l}^{\mathcal{T}_{2}},\cdots,\mathbf{z}_{i,l}^{\mathcal{T}_{V}} \right ]$. Finally, given that anomalous nodes typically do not follow temporal translation invariance, we hypothesize that anomalous nodes have low context consistency of their representations between timespan views, while normal nodes tend to have high consistency of their representations between timespan views. Our anomaly discriminator computes the anomaly score by capturing the context consistency and stability between timespan views:
\begin{equation}
\overline{S_{i}}=\sum_{q=1}^{V}\sum_{k\neq q}^{V}D\left ( \mathbf{z}_{i,l}^{\mathcal{T}_{q}},\mathbf{z}_{i,l}^{\mathcal{T}_{k}} \right )/\left ( V\cdot \left ( V-1 \right ) \right ),
\end{equation}

\begin{equation}
S\left ( v_{i} \right )= \overline{S_{i}}+\sqrt{\sum_{q=1}^{V}\sum_{k\neq q}^{V}\left ( D\left ( \mathbf{z}_{i,l}^{\mathcal{T}_{q}},\mathbf{z}_{i,l}^{\mathcal{T}_{k}} \right )-\overline{S_{i}} \right )^{2}/\left ( V\cdot \left ( V-1 \right ) \right )},
\label{eq:ad}
\end{equation}
where $S\left ( v_{i} \right )$ is the anomaly score of node $i$. The larger $S\left ( \cdot \right )$ means a higher abnormality. $D\left ( q,k \right )=1-q^{\top }k/\left\| q\right\|\left\| k\right\|$ is the consistency distance measure function. $\overline{S_{i}}$ computes the context consistency between timespan views, while Eq.~\ref{eq:ad} computes the stability between contexts.

\subsection{Complexity Analysis}
Our proposed method is agnostic to the size of the graph, with its parameters primarily concentrated in the base encoder and projection head, rendering it inherently lightweight. The time complexity is $O\left(L\left|\mathcal{V}\right|d^{2}+L\left| \mathcal{E}\right|d\right)$, where $\mathcal{V}$ is the set of nodes, $\mathcal{E}$ is the set of edges, $L$ represents the number of layers of the graph convolution, and $d$ is the node feature dimension. This is significantly lower than existing dynamic graph representation learning methods that additionally consider sequential models. Furthermore, the space complexity of our method's parameters is $O\left(Ld^{2}\right)$. However, existing dynamic graph representation learning methods typically integrate node and edge features into the model parameters, increasing the space complexity by $O\left(\left|\mathcal{V}\right|d+ \left|\mathcal{E}\right|d\right)$. More importantly, this would result in parameter space and runtime increasing significantly with the scale of the graph, rendering it impractical for large-scale graphs.

\subsection{Advantages over Dynamic Graph Methods}
Compared with previous dynamic graph work, CLDG and CLDG++ have the following advantages:

\textbf{Better Generality}. Existing dynamic graph work is generally designed solely for discrete-time dynamic graphs or continuous-time dynamic graphs. However, the timespan view sampling layer enables CLDG and CLDG++ to handle both discrete-time and continuous-time dynamic graphs.

\textbf{Better Scalability}. The encoder module of CLDG and CLDG++ can use any network architecture, and any effective encoder in the future can also be integrated into CLDG and CLDG++ in a way similar to hot swap. 

\textbf{Lower Space and Time Complexity}. A generalized dynamic graph learning paradigm is to use graph neural networks to model structural information, and then use sequential models to explicitly model evolutionary information. 
However, CLDG and CLDG++ implicitly exploit the additional temporal cues provided by dynamic graphs through contrastive learning, omitting the sequential model architecture. Therefore it has lower space and time complexity.

\section{Experiments}
In this section, we first introduce seven real-world dynamic graph datasets. Then, we conduct extensive experiments to prove the CLDG and CLDG++ yield consistent and significant improvements over baselines. Finally, ablation experiments on different components such as the timespan view sampling layer and base encoder layer. Meanwhile, the analysis of space and time complexity further proves the lightweight of our model. The code and data are publicly available \footnote{\url{https://github.com/yimingxu24/CLDG}}.
The experiment aims to answer the following questions:

$\bullet$
$\mathbf{RQ_{1}}$: Does our model yield state-of-the-art results for unsupervised dynamic graph representation learning in the node classification task?

$\bullet$
$\mathbf{RQ_{2}}$: Does our model achieve state-of-the-art results on the anomaly detection tasks?

$\bullet$
$\mathbf{RQ_{3}}$: How to choose the appropriate timespan as a contrastive view?

$\bullet$
$\mathbf{RQ_{4}}$: Does our model allow various encoder architecture choices without any limitations? Do different encoders have any effect on CLDG and CLDG++?

$\bullet$
$\mathbf{RQ_{5}}$: How about the time complexity and space complexity of our model?

$\bullet$
$\mathbf{RQ_{6}}$: How do different contrast pairs affect CLDG++?

$\bullet$
$\mathbf{RQ_{7}}$: How do various hyperparameters impact CLDG++ performance? 

\begin{table}[t]\small
\renewcommand\arraystretch{1.2}
\setlength\tabcolsep{15pt}
\centering
  \caption{Statistics of the Datasets.}
  \label{tab:SD}
  \begin{tabular}{cccccccc}
    \toprule
    % Dataset & node type & node & edge type & edge & attribute & label node\\
    \# Dataset & \# nodes & \# temporal edges & \# classes \\
    \midrule
    DBLP & 25,387 & 185,480 & 10 \\
    Bitcoinotc & 5,881 & 35,592 & 3 \\
    TAX & 27,097 & 315,478 & 12 \\
    BITotc & 4,863 & 28,473 & 7 \\
    BITalpha & 3,219 & 19,364 & 7 \\
    TAX51 & 132,524 & 467,279 & 51 \\
    Reddit & 898,194 & 2,575,464 & 3 \\
    \bottomrule
  \end{tabular}
\end{table}

\subsection{Experiment Preparation}
\subsubsection{Real-world Dataset}\label{dataset}
Seven datasets from four fields (i.e. academic citation network, tax transaction network, Bitcoin network, and Reddit hyperlink network) are used to evaluate the quality of representations learned by CLDG and CLDG++. The statistics of the seven datasets are shown in Table~\ref{tab:SD}.

% $\bullet$
\textbf{Academic Citation Network}. The DBLP dataset is extracted from the DBLP \footnote{\url{http://dblp.uni-trier.de}}. The label is the research field of researchers. The label information divides researchers into ten different research areas, and each researcher has only one label. 

\textbf{TAX Transaction Networks}. The TAX and TAX51 company industry classification datasets are extracted from the tax transaction network data of two cities provided by the tax bureau. It consists of companies as vertices, and transaction relationships between companies as edges. The industry code of each company is provided by the tax bureau as labels of the dataset.

\textbf{Bitcoin Networks}. Bitcoin is a cryptocurrency used for anonymous transactions on the web. Due to the risk of trading with anonymity, this has led to the emergence of several exchanges where Bitcoin users rate how much they trust other users. Bitcoinotc, BITotc~\citep{kumar2016edge, liu2021inductive} and BITalpha~\citep{kumar2016edge, liu2021inductive} are three datasets from two bitcoin trading platforms, OTC and alpha~\footnote{\url{https://www.bitcoin-otc.com/} and \url{https://btc-alpha.com/}}. The label is the credit rating of the Bitcoin user.

\textbf{Reddit Hyperlink Network}. 
We construct the Reddit dataset from the Subreddit Hyperlink Network~\footnote{\url{http://snap.stanford.edu/data/soc-RedditHyperlinks.html}}. It consists of posts and hyperlinks in the community as nodes. The source and link relationship between each hyperlink and the post as edges. Each hyperlink is annotated with the sentiment of the source community post towards the target community post. This dataset can be used for sentiment classification.

For the anomaly detection task, since the above datasets have no anomalies by default, we follow previous work~\citep{ding2019interactive,song2007conditional} and inject a combined set of structural and attribute anomalies for each dataset. Specifically, for structural anomalies, we randomly select some nodes that may be initially unconnected, and add links with the same timestamps between them, making them a fully connected clique. The timestamp of each clique is randomly generated within the overall timespan of the graph. All nodes in such a clique are marked as structural anomalies because the nodes within the clique are much more closely connected to each other than the average, which can be regarded as a typical structural anomaly in the real world~\citep{skillicorn2007detecting,liu2021anomaly}. Meanwhile, for attribute anomalies, we create feature anomalies by perturbing the node attributes of a timespan. When injecting a single node feature anomaly, we randomly pick a node $v_i$ as the target node and sample another $k$ nodes as the candidate set. We choose the node $v_j$ with the largest Euclidean distance from the feature of node $v_i$ among the $k$ nodes. Then, we inject the feature of $v_j$ into a certain timespan of $v_i$, while leaving the feature of $v_i$ unchanged in other timespans. To ensure a sufficiently large disturbance amplitude, we set $k=50$. Since the static graph method cannot handle temporal inputs, we introduce the average of all timespan node features to construct feature anomalies for the static graph method. Finally, we injected 600, 300, 600, 300, 300, 6000, and 6000 anomalies into 7 datasets respectively.

\subsubsection{Evaluation Protocols}
We divide the seven datasets according to the 1:1:8 split method of training set: validation set: test set.
Semi-supervised methods train the model through training and validation sets, outputting predicted labels for each node in the test set. 
Our method and other unsupervised methods follow a linear evaluation scheme as introduced in~\citep{velickovic2019deep}, where each model is firstly trained in an unsupervised manner. After training, freeze the model parameters and output the learned representations for all nodes. Subsequently, the representation is fed into a linear classifier, noting that only the node embeddings in the training set are used to train the classifier. We report the Accuracy and Weighted-F1 metrics for node classification in the test set.
For the anomaly detection task, we still measure our proposed framework and baselines based on the ROC-AUC (AUC for short), which is widely used to evaluate anomaly detection performance in previous work~\citep{ding2019deep,peng2018anomalous}.

\subsubsection{Implementation Details}
Adam optimizer~\citep{kingma2014adam} is used in the training model stage and the training linear classifier stage. In the training model stage, the learning rate of the encoder and projection head is $4 \times 10^{-3}$, and the weight decay is $5 \times 10^{-4} $. In the training linear classifier stage, the learning rate is $10^{-2} $, the weight decay is $ 10^{-4} $. The base encoder adopts a two-layer GCN, the batch size is 256, the hidden layer dimension is 128, and the output dimension is 64. The random sampling strategy is chosen for the BITotc and TAX51 datasets, while the sequential sampling strategy is employed for the other datasets. The heat kernel is selected as the graph diffusion instance for the Bitcoinotc and TAX51 datasets, and others are PPR. 

\begin{table*}[]\small
\centering
\renewcommand\arraystretch{1.24}
\setlength\tabcolsep{2pt}
\caption{Experimental results (\%) of classification tasks on seven datasets. We report both mean Accuracy and Weighted-F1, the input column highlights the data required for model training ($\mathbf{X}$ is the node features, $\mathbf{A}$ is the adjacency matrix, $\mathbf{S}$ is the diffusion matrix, $\mathbf{T}$ is the time information, and $\mathbf{Y}$ is the node labels). Bold represents the optimal in the unsupervised method, and underlined represents the global optimal.}
\resizebox{\textwidth}{!}{
\begin{tabular}{ccccccccccccccccc}
\hline \hline
                              & \multirow{2}{*}{Method} & \multirow{2}{*}{Input} & \multicolumn{2}{c}{DBLP} & \multicolumn{2}{c}{Bitcoinotc} & \multicolumn{2}{c}{TAX} & \multicolumn{2}{c}{BITotc} & \multicolumn{2}{c}{BITalpha} & \multicolumn{2}{c}{TAX51} & \multicolumn{2}{c}{Reddit} \\
                              \cline{4-17}
                              &                         &                        & Acc     & Wei     & Acc        & Wei        & Acc     & Wei    & Acc      & Wei      & Acc       & Wei & Acc       & Wei & Acc       & Wei       \\ \hline
\multirow{4}{*}{\rotatebox{90}{Supervised}}   & LP                      & $\mathbf{A,Y}$     & 52.18     & 51.54        & 41.97        & 31.25           & 35.67     & 30.88       & 60.24     & 50.28         & 76.97       & 67.18     & 28.82       & 24.36     & 66.71       & 60.88          \\
                              & GCN                     & $\mathbf{X, A, Y}$                & 71.35     & 71.08        & 54.61        & 54.41           & 71.65     & 71.37        & 59.24      & 50.52         & 76.19       & 73.65 & 40.42       & 33.64      & 67.83         & 63.26          \\
                              & GAT                     & $\mathbf{X, A, Y}$                & 70.01     & 69.21        & 52.46        & 50.60           & 62.00      & 59.78       & 62.59      & 48.34         & 80.21       & 71.40         & 38.82       & 30.21         & 69.31       & 58.01          \\
                              & GraphSAGE               & $\mathbf{X, A, Y}$                & 72.36     & 71.99        & 57.29        & 56.30           & 64.36     & 63.73       & 62.68      & $\underline{56.99}$         & 79.89       & $\underline{74.24}$         & 40.80       & $\underline{33.79}$     & 71.37       & \underline{65.74}          \\ \hline
\multirow{10}{*}{\rotatebox{90}{Unsupervised}} & CAW
                     & $\mathbf{X, A, T}$                & 55.64     & 51.38        & 59.85        & 57.92           & 53.25     & 47.11       & 63.56      & 53.59         & 77.64       & 71.99         & 36.52       & 29.52         & 67.79       & 57.80           \\
                              & TGAT                    & $\mathbf{X, A, T}$                & 57.48     & 51.96        & 58.56        & 55.73           & 50.12     & 45.62       & 62.53      & 49.53         & 77.63       & 68.03         & 34.50       & 28.44         & 65.57       & 56.81          \\
                              & DySAT                  & $\mathbf{X, A, T}$                & 54.12     & 52.48        & 51.32        & 48.80            & 52.31     & 51.42       & 62.01      & 48.70          & 77.61       & 68.07          & 23.81       & 22.89           & -     & -          \\
                              & MNCI                    & $\mathbf{X, A, T}$                & 67.03     & 66.46        & 55.14        & 54.79           & 45.44     & 37.13       & 63.49      & 54.66         & 79.04       & 71.80         & 39.13       & 32.29         & 70.94       & $\mathbf{65.26}$           \\
                              & DGI                 & $\mathbf{X, A}$                & 69.97     & 69.40        & 56.67        & 55.34           & 66.57     & 65.87       & 61.68      & 51.65         & 78.56       & $\mathbf{73.19}$     & 39.20       & 31.61     & 70.58       & 60.95          \\
                              & GRACE                 & $\mathbf{X, A}$                & 71.27     & 70.71        & 54.50        & 53.48           & 67.43     & 66.79       & 62.49      & 48.06         & 77.60       & 67.81           & -     & -           & -     & -          \\
                              & MVGRL                   & $\mathbf{X, A, S}$                & 71.32     & 70.91        & 55.31        & 54.54            & 67.70     & 67.08       & 59.62      & 53.00          & 75.26       & 72.71           & -     & -           & -     & -          \\
                              & CCA-SSG                 & $\mathbf{X, A}$                & 68.28     & 67.60        & 56.48        & 54.83           & 67.62     & 67.14       & 63.47      & 51.70         & 79.97       & 72.69         & 37.04       & 29.69         & 69.46       & 58.69           \\ 
                              & $\mathbf{CLDG}$                    & $\mathbf{X, A, T}$                & 71.80     & 71.55        & 59.17        & 58.45           & 69.62     & 69.18       & $\underline{\mathbf{65.68}}$      & $\mathbf{54.74}$         & 80.61       & 72.90         & 40.44       & 32.36       & 71.69      & 62.87 \\          
                              & $\mathbf{CLDG++}$                 & $\mathbf{X, A, S, T}$                & $\underline{\mathbf{72.94}}$     & $\underline{\mathbf{72.69}}$        & $\underline{\mathbf{59.88}}$        & $\underline{\mathbf{58.96}}$           & $\underline{\mathbf{73.21}}$     & $\underline{\mathbf{73.01}}$       & 65.37      & 54.44         & $\underline{\mathbf{80.63}}$       & 72.87         & $\underline{\mathbf{41.05}}$       & $\mathbf{33.15}$         & $\underline{\mathbf{71.73}}$       & 62.56  
                              \\ \hline
                              \hline
\end{tabular}}
\label{tab:nc}
\end{table*}

\subsection{Node Classification ($\mathbf{RQ_{1}}$)}
To answer $\mathbf{RQ_{1}}$, we compare CLDG and CLDG++ with 12 state-of-the-art algorithms on the seven dynamic graph datasets. 
The semi-supervised model in the baseline includes label propagation (LP), GCN, GAT, and GraphSAGE. We also compare with unsupervised dynamic graph models including CAW, TGAT, DySAT, and MNCI, and unsupervised contrastive learning methods including DGI, GRACE, MVGRL, and CCA-SSG. We use the Accuracy and Weighted-F1 as the evaluation metrics, bold indicates that the method is optimal among unsupervised methods, and underlined indicates that it is optimal among all baselines. We report classification results on seven datasets in Table~\ref{tab:nc}. 

We can observe that both CLDG and CLDG++ outperform other unsupervised methods. Among the Accuracy and Weighted-F1 metrics of the seven datasets, 12 metrics are optimal among the unsupervised methods. CLDG outperforms the previous state-of-the-art GraphSAGE model by 1.47\% on the average of all metrics. CLDG++, which introduces local-global and global-global contrasts, outperforms CLDG in 10 metrics, which indicates that retaining more information helps learn high-quality node representations. Supervised methods are overall very competitive. In contrast, LP limits expressive power by ignoring the node feature matrix $\mathbf{X}$. However, without label supervision, our method even shows superior results than all supervised models in 10 metrics. This proves that using the local-local and global-global contrasts in different timespan views, as well as the local-global contrast in the same timespan view in dynamic graphs, provides rich graph information and is sufficient for classification tasks.

In summary, experimental results demonstrate that CLDG++ generalizes contrastive learning from static graphs to dynamic graphs through multi-scale temporal translation invariance on graphs, and implicitly utilizes temporal information to achieve state-of-the-art results.

\begin{table*}[]\small
% \begin{table*}[]
\centering
\renewcommand\arraystretch{1.3}
\caption{AUC Scores (\%) for Anomaly Detection on seven datasets (OOM: CPU/CUDA Out of Memory). Bold represents the global optimal.}
\resizebox{\textwidth}{!}{
\begin{tabular}{ccccccccc}
\hline \hline
& \multirow{1}{*}{Method} & \multicolumn{1}{c}{DBLP} & \multicolumn{1}{c}{Bitcoinotc} & \multicolumn{1}{c}{TAX} & \multicolumn{1}{c}{BITotc} & \multicolumn{1}{c}{BITalpha} & \multicolumn{1}{c}{TAX51} & \multicolumn{1}{c}{Reddit} \\ \hline
% \cline{4-17}

& AEGIS 
 & $59.18_{\pm 3.17}$     & $50.97_{\pm 6.15}$        & $59.62_{\pm 4.06}$        & $53.17_{\pm 0.57}$           & $54.13_{\pm 4.57}$     & $54.52_{\pm 1.67}$       & OOM  \\
& CoLA   & $71.06_{\pm 0.24}$     & $79.44_{\pm 1.46}$        & $66.68_{\pm 0.24}$        & $81.80_{\pm 0.31}$           & $77.65_{\pm 0.19}$     & $77.29_{\pm 0.12}$       & OOM \\ 
& ANEMONE    & $70.69_{\pm 1.00}$     & $79.28_{\pm 0.56}$        & $67.46_{\pm 1.29}$        & $80.09_{\pm 0.93}$           & $72.70_{\pm 1.47}$     & $76.61_{\pm 0.23}$       & OOM \\ 
& GRADATE    & $69.53_{\pm 0.10}$     & $74.14_{\pm 0.32}$        & $70.62_{\pm 0.49}$        & $73.12_{\pm 0.42}$           & $68.64_{\pm 0.43}$     & OOM       & OOM \\
& NetWalk    & $54.42_{\pm 0.33}$     & $56.13_{\pm 0.71}$        & $55.42_{\pm 0.64}$        & $58.46_{\pm 0.44}$           & $64.12_{\pm 0.60}$     & OOM       & OOM \\ 
& TADDY    & $50.93_{\pm 0.59}$     & $50.58_{\pm 0.33}$        & $50.83_{\pm 0.55}$        & $51.48_{\pm 0.38}$           & $50.42_{\pm 0.30}$     & OOM       & OOM \\ 
& SAD    & $78.24_{\pm 0.31}$     & $75.67_{\pm 0.28}$        & $70.61_{\pm 0.52}$        & $71.29_{\pm 0.46}$           & $71.83_{\pm 0.45}$     & $61.70_{\pm 0.79}$       & $72.19_{\pm 0.60}$ \\ 
& $\mathbf{CLDG}$   &  $84.89_{\pm 0.24}$     & $80.21_{\pm 0.99}$        & $71.81_{\pm 0.15}$        & $81.55_{\pm 0.84}$           & $77.70_{\pm 0.99}$     & $79.51_{\pm 0.13}$       & $71.53_{\pm 0.28}$ \\ 
& $\mathbf{CLDG++}$     &  $\mathbf{86.41}_{\pm 0.32}$     & $\mathbf{81.97}_{\pm 1.08}$        & $\mathbf{72.86}_{\pm 0.28}$        & $\mathbf{82.92}_{\pm 0.54}$           & $\mathbf{79.71}_{\pm 0.36}$     & $\mathbf{81.02}_{\pm 0.26}$       & $\mathbf{72.77}_{\pm 0.08}$ 
\\ \hline
\hline
\end{tabular}}
\label{tab:ad}
\end{table*}

\subsection{Anomaly Detection ($\mathbf{RQ_{2}}$)}
To address $\mathbf{RQ_{2}}$, we compare our model with seven popular graph-based anomaly detection methods chosen as baselines. The unsupervised models in the baselines include AEGIS, ANEMONE, CoLA, GRADATE, NetWalk, and TADDY. SAD represents a semi-supervised anomaly detection approach. Among the baselines, models NetWalk, TADDY, and SAD incorporate temporal information in their methods. NetWalk and TADDY primarily focus on edge anomaly detection. Therefore, after training the models, we preserve the learned node representations and uniformly input them into our proposed anomaly discriminator to compute node anomaly scores. Table~\ref{tab:ad} reports the anomaly detection performance of all methods on 7 datasets. 
First, CLDG outperforms all unsupervised baselines, while CLDG++ surpasses all baselines, which strongly proves the success of CLDG and its extended version in the anomaly detection task. Second, we can observe that the three contrastive learning-based baselines CoLA, ANEMONE and GRADATE are all very competitive, indicating that contrastive learning is generally effective on anomaly detection tasks. Third, one possible reason for the mediocre performance of NetWalk and TADDY on several datasets is that they were originally designed to detect abnormal edges. SAD is specifically designed to detect node anomalies while using a semi-supervised training paradigm to achieve better performance. Finally, the above baselines typically require complete adjacency and attribute matrices as input, making them impractical for large-scale graph data due to the explosive memory requirements.

\begin{figure*}[!ht]
\centering
\subfloat[BITalpha]{\label{fig:vis1}\includegraphics[width=0.45\linewidth]{
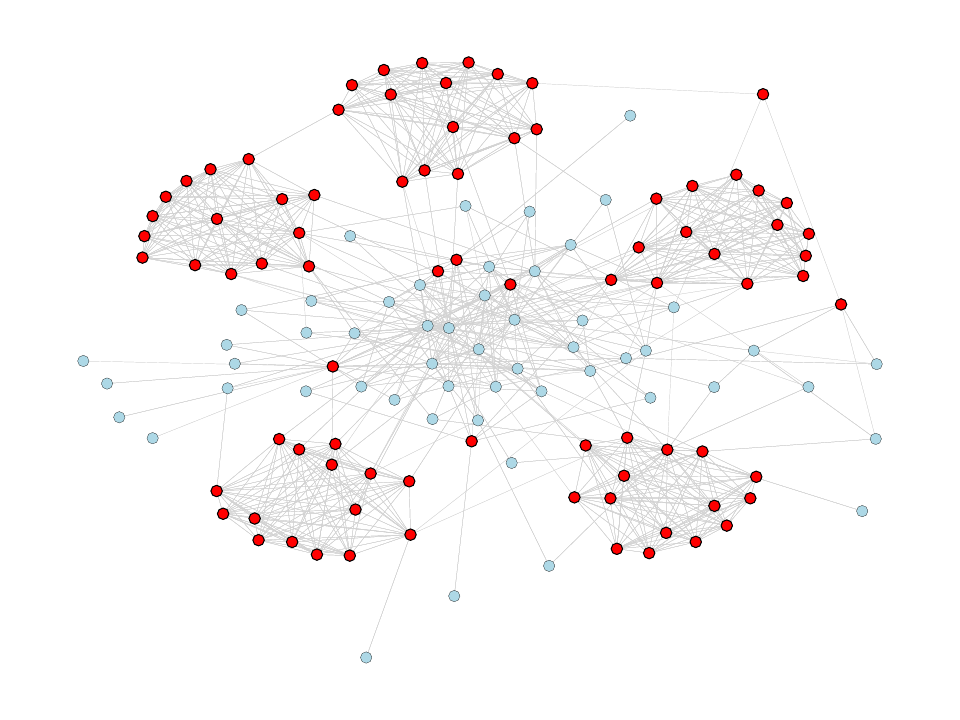}}
\subfloat[CoLA]{\label{fig:vis2}\includegraphics[width=0.45\linewidth]{
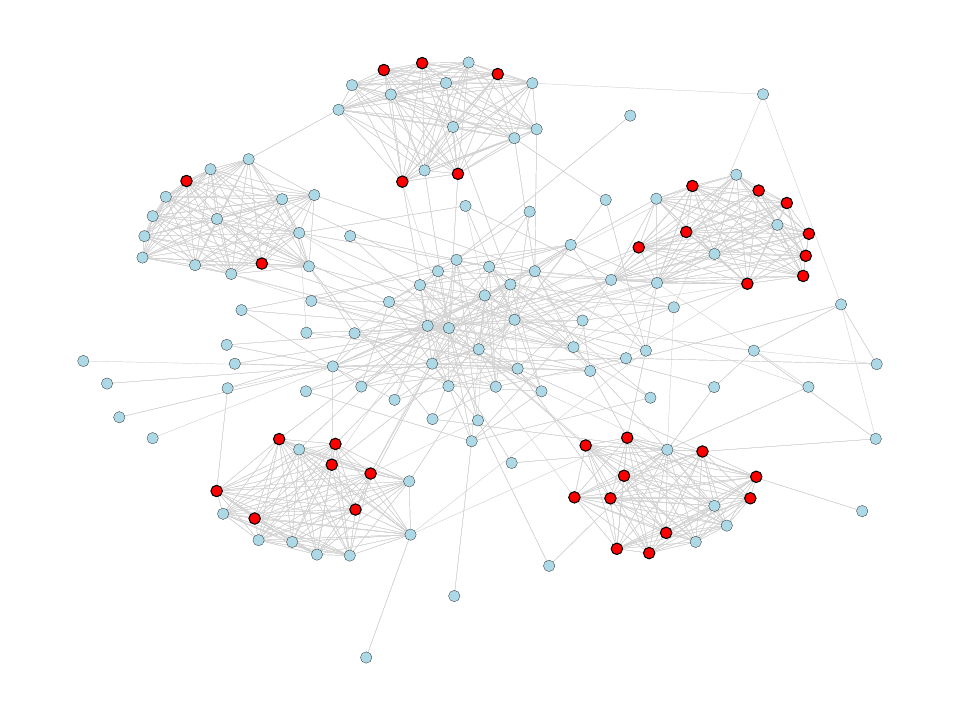}}\\
\centering
\subfloat[CLDG]{\label{fig:vis3}\includegraphics[width=0.45\linewidth]{
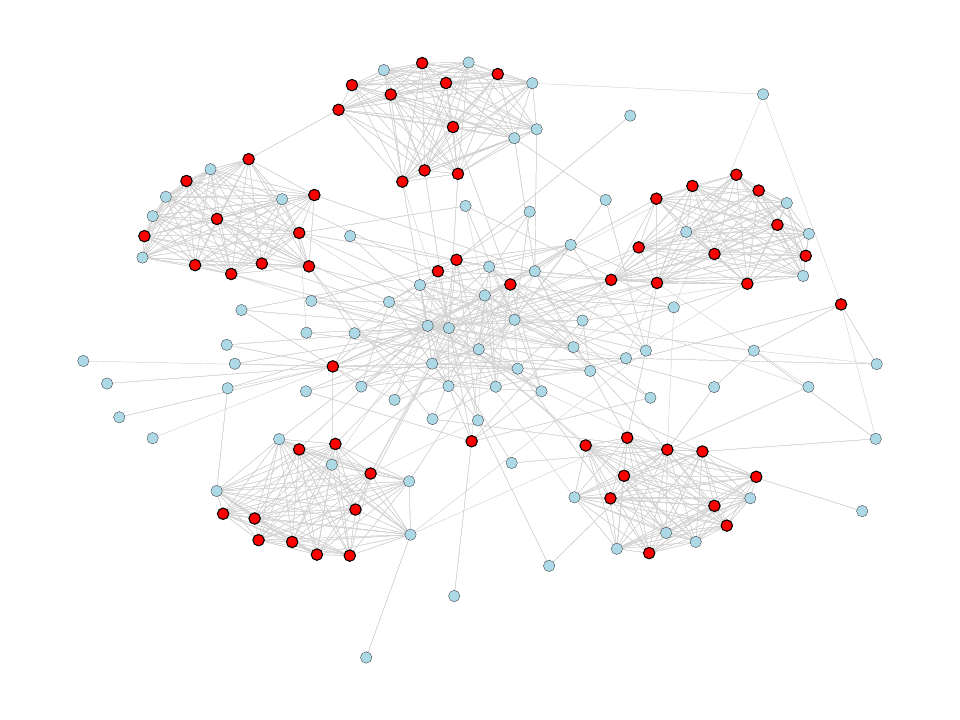}}
\subfloat[CLDG++]{\label{fig:vis4}\includegraphics[width=0.45\linewidth]{
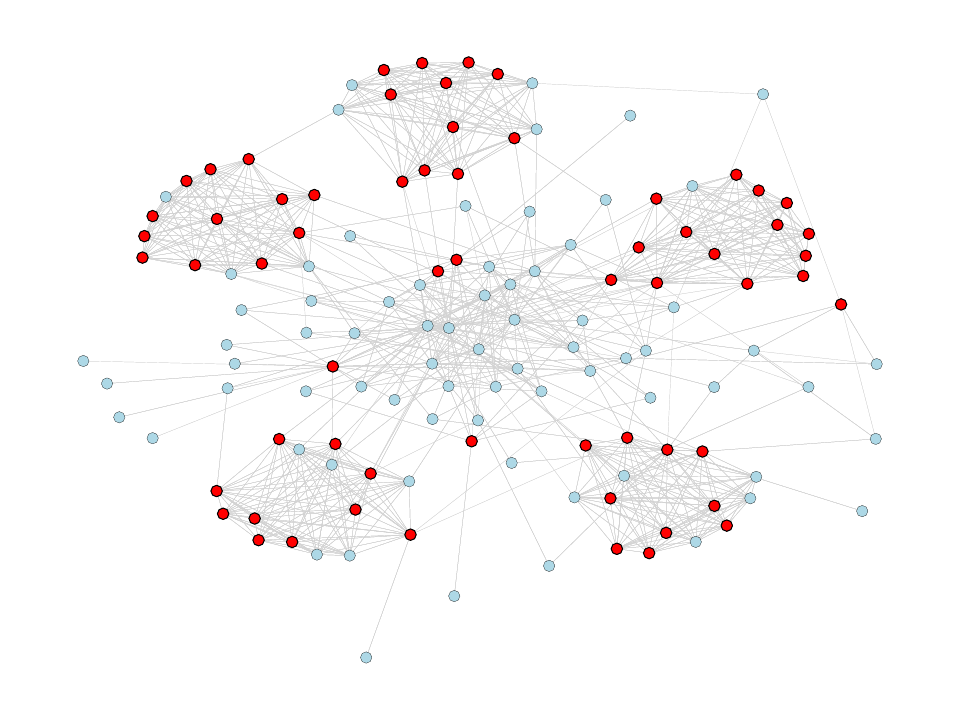}}\\	
\caption{Visualization of detection results on BITalpha dataset. Abnormal/normal nodes or nodes detected as abnormal/normal by the model are marked in red/blue. (a) is the ground truth of the sampled BITalpha dataset, and the others are the anomaly detection results of CoLA, CLDG, and CLDG++ respectively.}
\label{fig:ad_vis}
\end{figure*}

In addition to the aforementioned quantitative analysis, we also present qualitative visual case studies. Figure~\ref{fig:vis1} illustrates a sampled subgraph from BITalpha, where the ground truth anomalies and normal nodes are denoted by red and blue colors, respectively. In CoLA, CLDG, and CLDG++, we mark the 300 nodes in the dataset with the highest node anomaly scores in red. As shown in Figure~\ref{fig:vis2}, such as the clique located in the upper-left corner, most of the structural anomalous nodes are misclassified as normal nodes by CoLA. The observations from Figure~\ref{fig:vis3} and Figure~\ref{fig:vis4} reveal that our model accurately detects the majority of abnormal nodes, including structural abnormalities in various clusters. Moreover, CLDG++ exhibits higher detection accuracy compared to CLDG. This finding underscores the strong anomaly detection performance of both CLDG and its extended version, aligning with the results presented in Table~\ref{tab:ad}.

\newcolumntype{x}[1]{>{\centering\arraybackslash}p{#1pt}}
\begin{table*}[t]\small
	\renewcommand\arraystretch{1.24}
	\caption{Comparison of different timespan view sampling architectures, sampling strategy, sampling view timespan size $s$ and number of sampling views $v$ in the DBLP and TAX dataset. The sampling strategy with a low overlap rate between views, such as sequential and random, and sampling more views are beneficial to CLDG, and the timespan of sampling views is robust to CLDG.}
	\label{tab:gs}
 
	\subfloat[ \footnotesize{Dataset: \textbf{DBLP}, base encoder: \textbf{GCN}, epoch: \textbf{200}}.]{
    \resizebox{\textwidth}{!}{
    \label{tab:gs_dblp}
    \begin{tabular}{x{14}|x{26}x{26}x{26}|x{26}x{26}x{26}|x{26}x{26}x{26}|x{26}x{26}x{26}}
		&  \multicolumn{3}{c|}{ \textbf{Sequential} } &  \multicolumn{3}{c|}{ \textbf{High Overlap Rate} }  &  \multicolumn{3}{c|}{ \textbf{Low Overlap Rate}} &  \multicolumn{3}{c}{ \textbf{Random}} \\
		\multicolumn{1}{c|}{\diagbox{v}{s}}  & 6 & 8 & 10 & 6 & 8 & 10 & 6 & 8 & 10 & 6 & 8 & 10  \\
		\shline
        2 & 71.11    & 71.01    & 70.84    & 69.44       & 68.92      & 69.16      & 70.26      & 70.59      & 70.10      & 71.01   & 70.69   & 70.48  \\
        3 & 71.34    & 71.10    & 71.17    & 69.92       & 69.78      & 69.56      & 70.85      & 70.73      & 70.79      & 70.95   & 71.06   & 70.75  \\
        4 & 71.40    & 71.34    & 71.00    & 70.01       & 69.75      & 69.82      & 70.80      & 70.60      & 70.73      & 70.53   & 71.06   & 70.61  \\
        5 & 71.55    & 71.49    & 71.23    & 70.48       & 70.62      & 70.23      & 71.16      & 70.96      & 70.54      & 71.04   & 70.77   & 70.74 \\
	\end{tabular}}}
	\hfill
	
	\subfloat[ \footnotesize{Dataset: \textbf{TAX}, base encoder: \textbf{GCN}, epoch: \textbf{200}}.]{
    \resizebox{\textwidth}{!}{
    \label{tab:gs_tax}
    \begin{tabular}{x{14}|x{26}x{26}x{26}|x{26}x{26}x{26}|x{26}x{26}x{26}|x{26}x{26}x{26}}
		&  \multicolumn{3}{c|}{ \textbf{Sequential} } &  \multicolumn{3}{c|}{ \textbf{High Overlap Rate} }  &  \multicolumn{3}{c|}{ \textbf{Low Overlap Rate}} &  \multicolumn{3}{c}{ \textbf{Random}} \\
		\multicolumn{1}{c|}{\diagbox{v}{s}}  & 6 & 8 & 10 & 6 & 8 & 10 & 6 & 8 & 10 & 6 & 8 & 10  \\
		\shline
        2 & 67.05    & 67.94    & 65.75    & 63.52       & 62.87      & 62.68      & 64.98      & 64.22      & 63.51      & 65.26   & 65.13   & 65.08  \\
        3 & 67.65    & 67.25    & 66.58    & 63.79       & 63.68      & 63.04      & 64.53      & 64.76      & 63.91      & 66.33   & 65.57   & 65.63  \\
        4 & 68.99    & 67.85    & 67.51    & 64.22       & 63.69      & 63.40      & 64.91      & 64.53      & 64.05      & 66.80   & 66.46   & 65.97  \\
        5 & 68.98    & 67.01    & 67.60    & 64.26       & 63.35      & 63.50      & 65.03      & 64.76      & 64.11      & 66.79   & 66.18   & 66.19
	\end{tabular}}}
	\hfill 
\end{table*}

\subsection{View Sampling Architecture ($\mathbf{RQ_{3}}$)}
The timespan view sampling module consists of three factors: sampling strategy, view timespan factor $s$ and the number of views sampled $v$. We design four different sampling strategies, sequential, high overlap rate, low overlap rate and random, and the details of the sampling strategy are demonstrated in this section~\ref{tpgs}. $v$ controls the number of the sampling timespan views, and $s$ controls the timespan size of the view.  
We show in Table~\ref{tab:gs}a and Table~\ref{tab:gs}b how the three factors jointly affect the performance of CLDG on the DBLP and TAX datasets.

The variation on the dynamic graph is continuous and smooth, and if the timespan views overlap physically, better results may be achieved by sharing more similar semantic contexts. Therefore, the main difference between the four sampling strategies is the overlap rate between views. However, the better performers of the four strategies are sequential and random. The high overlap rate strategy tends to be the worst performer, on average 1.41\% and 4.01\% lower than the sequential strategy in the DBLP and TAX datasets, respectively. This is a very interesting phenomenon, illustrating that on a dynamic graph with temporal translation invariance, a high overlap rate may lead to an overly simplistic contrastive learning task, thereby compromising the robustness of the model. The sequential and randomized approaches maintain high-level semantic consistency rather than just low-level physical consistency allowing for better performance on the test set. The view timespan factor $s$, i.e., constructed views with different timespans, is relatively robust in both datasets and seems to have little effect on CLDG. In practice, we can choose some timespans with physical significance, such as day, week, month, and so on. Finally, on the number of timespan views v, we find that $v = 3,4,5$ on average 0.37\%, 0.34\%, 0.60\%, and 0.40\%, 0.87\%, 0.82\% higher than $v=2$, in DBLP and TAX datasets, respectively. This indicates that more timespan views are beneficial for CLDG, but the average improvement is smaller after the number of views exceeds 3.

In general, the sampling strategy with the low overlap rate between views and sampling more views is beneficial to CLDG, and the timespan of constructed views is robust.

\begin{table}[]\footnotesize 
\centering
\setlength{\belowcaptionskip}{0.05cm}  
\renewcommand\arraystretch{1.3}
\setlength\tabcolsep{6.6pt}
\caption{Scalability verification of CLDG and CLDG++ by comparing the results of different encoder variants on three datasets.}
\begin{tabular}{cccccccccc}
\hline
\hline
  & \multirow{2}{*}{Encoder} & \multicolumn{2}{c}{DBLP} & \multicolumn{2}{c}{BITalpha} & \multicolumn{2}{c}{TAX51} \\
  \cline{3-8}
  &                         & Acc     & Wei-F1     & Acc        & Wei-F1        & Acc     & Wei-F1       \\ \hline
    \multirow{3}{*}{\rotatebox{90}{CLDG}}   & GCN       & 71.48 & 71.18 & 80.14 & 72.18 & 40.07 & 32.02 \\
    & GAT       & 73.49 & 73.16 & 80.12 & 71.79 & 41.64 & 32.85 \\
    & GraphSAGE & 73.19 & 72.51 & 80.09 & 71.97 & 41.34 & 34.78 \\ \hline
    \multirow{3}{*}{\rotatebox{90}{CLDG++}}   & GCN       & 72.98 & 72.72 & 80.14 & 72.27 & 40.68 & 32.58 \\
    & GAT       & 74.47 & 74.24 & 80.21 & 71.93 & 42.96 & 34.51 \\
    & GraphSAGE & 73.71 & 73.06 & 80.19 & 71.68 & 43.80 & 34.85 \\ \hline
    \hline

\end{tabular}
\label{tab:encoder}
\end{table}

\subsection{Encoder Architecture ($\mathbf{RQ_{4}}$)}
To answer $\mathbf{RQ_{4}}$, we implement three classical GNN encoders (GCN, GAT and GraphSAGE) for CLDG and CLDG++. We conduct experiments on the DBLP, BITalpha, and TAX51 datasets. The number of layers of the encoder is 2, the timespan view sampling strategy is sequential, $s$ is 4, $v$ is 2, the epoch is 200, 100 and 200, respectively. Table~\ref{tab:encoder} shows the classification results of different encoder architectures in the test set. The experimental results show that all three encoders have very competitive performance. 
Under the current parameter settings, GraphSAGE performs best in CLDG, averaging 1.14\% higher than CLDG with the GCN encoder on the three datasets. In CLDG++, GAT is the best encoder choice, 1.16\% higher than GCN. Meanwhile, we can observe that CLDG++, which considers global information, shows better performance compared to CLDG in various encoder selections, improving by 0.72\%, 0.88\%, and 0.57\% in GCN, GAT, and GraphSAGE respectively.

In conclusion, our experiments illustrate that CLDG and CLDG++ exhibit flexibility in choosing various encoder architectures without any limitations. Notably, all employed encoders consistently achieve highly competitive results, showcasing the remarkable robustness of CLDG and CLDG++.

\subsection{Space and Time Complexity Analysis ($\mathbf{RQ_{5}}$)}
To answer $\mathbf{RQ_{5}}$, we compared CLDG and CLDG++ with four other dynamic graph methods (CAW, TGAT, DySAT, and MNCI) in terms of space and time complexity. In Table~\ref{tab:STCA}, we measure the space and time complexity by the number of parameters and the model training time, respectively. In terms of space complexity, first, the parameters of our model do not increase with the increase in graph size. However, other dynamic graph models have a maximum of 104.21, 123.54, 1.00, and 265.39 times increase in parameters on the seven datasets, respectively. The CLDG and CLDG++ parameters are derived from the base encoder and the projection head, and the base encoder can also be replaced with an encoder of lower space complexity and time complexity in the future. Then, the parameters of CLDG are much smaller than the existing SOTA dynamic graph model. The parameters of CAW, TGAT, DySAT, and MNCI are 17,840.6, 17,813.8, 2.0, and 5,785.4 times of CLDG on the Reddit dataset, respectively. Finally, in terms of time complexity, CLDG is implemented with a graph neighbor sampler, and the running time of CLDG is 946.4, 158.2, 132.7, and 30.1 times faster than CAW, TGAT, DySAT, and MNCI on TAX51 dataset. Due to the graph diffusion performed on the adjacency matrix of each sampled timespan view in CLDG++, the running time of the method increases. However, we still observe that the running time of CLDG++ remains significantly lower compared to existing dynamic graph representation learning methods.

In conclusion, CLDG and CLDG++ have lower space and time complexity than existing dynamic graph methods, and are more significant on larger-scale graphs. It is proved that our method is easier to generalize to larger-scale graph learning.

\begin{table*}[]
\centering
\renewcommand\arraystretch{1.3}
\setlength\tabcolsep{3pt}
\caption{Comparison of space and time complexity of dynamic graph method on seven datasets. CLDG has lower space and time complexity.}
\resizebox{\textwidth}{!}{
\begin{tabular}{ccccccccccccccccc}
\hline
\hline
  & \multirow{2}{*}{Method} & \multicolumn{2}{c}{DBLP} & \multicolumn{2}{c}{Bitcoinotc} & \multicolumn{2}{c}{TAX} & \multicolumn{2}{c}{BITotc} & \multicolumn{2}{c}{BITalpha} & \multicolumn{2}{c}{TAX51} & \multicolumn{2}{c}{Reddit}\\
  \cline{3-16}
  &                         & Param     & Time     & Param        & Time        & Param     & Time        & Param     & Time        & Param     & Time        & Param     & Time        & Param     & Time       \\ \hline
    & CAW       & 56.76M & 9403s & 13.39M & 3498s & 90.47M & 30702s & 11.31M & 2510s & 8.56M & 1768s & 156.32M & 329336s & 892.03M & 361580s \\
    & TGAT       & 55.41M & 5607s & 12.05M & 669s & 89.13M & 6067s & 9.96M & 1713s & 7.21M & 642s & 154.98M & 55066s & 890.69M & 50441s \\
    & DySAT & 0.10M & 5103s & 0.10M & 513s & 0.10M & 5781s & 0.10M & 422s & 0.10M & 456s & 0.10M & 46173s & 0.10M & - \\ 
    & MNCI & 8.22M & 4308s & 1.94M & 745s & 8.78M & 7580s & 1.62M & 550s & 1.09M & 481s & 42.72M & 10463s & 289.27M & 46586s \\ 
    & CLDG & \textbf{0.05M} & \textbf{297s} & \textbf{0.05M} & \textbf{136s} & \textbf{0.05M} & \textbf{312s} & \textbf{0.05M} & \textbf{74s} & \textbf{0.05M} & \textbf{248s}  & \textbf{0.05M} & \textbf{348s} & \textbf{0.05M} & \textbf{551s}\\
     & CLDG++ & \textbf{0.06M} & \textbf{3989s} & \textbf{0.06M} & \textbf{280s} & \textbf{0.06M} & \textbf{923s} & \textbf{0.06M} & \textbf{257s} & \textbf{0.06M} & \textbf{389s}  & \textbf{0.06M} & \textbf{1728s} & \textbf{0.06M} & \textbf{3865s}\\ \hline
    \hline

\end{tabular}}
\label{tab:STCA}
\end{table*}

\begin{figure*}
\centering
\setlength{\belowcaptionskip}{-0.3cm}
\subfloat[]{\label{fig:abl1}\includegraphics[width=0.5\linewidth]{
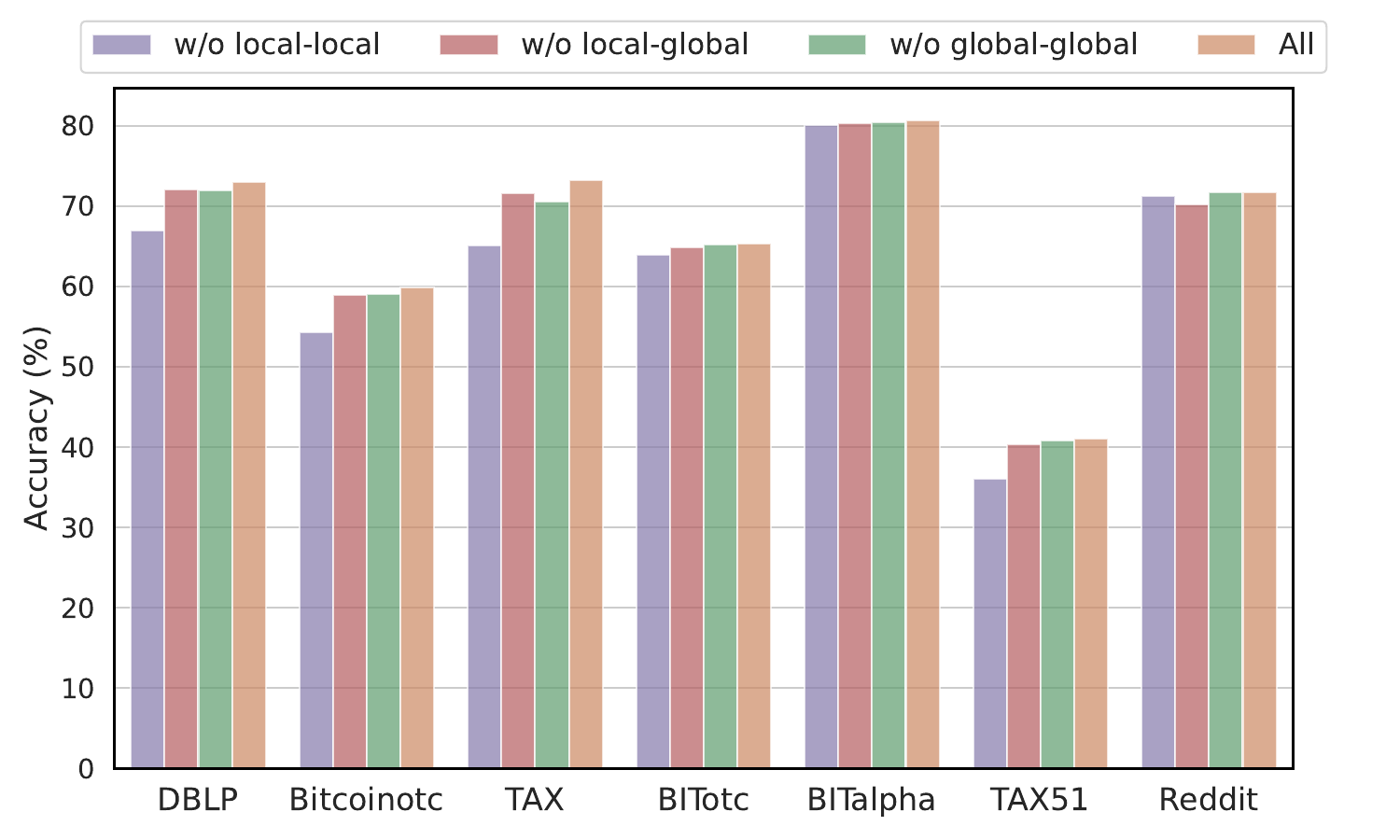}}
\subfloat[]{\label{fig:abl2}\includegraphics[width=0.5\linewidth]{
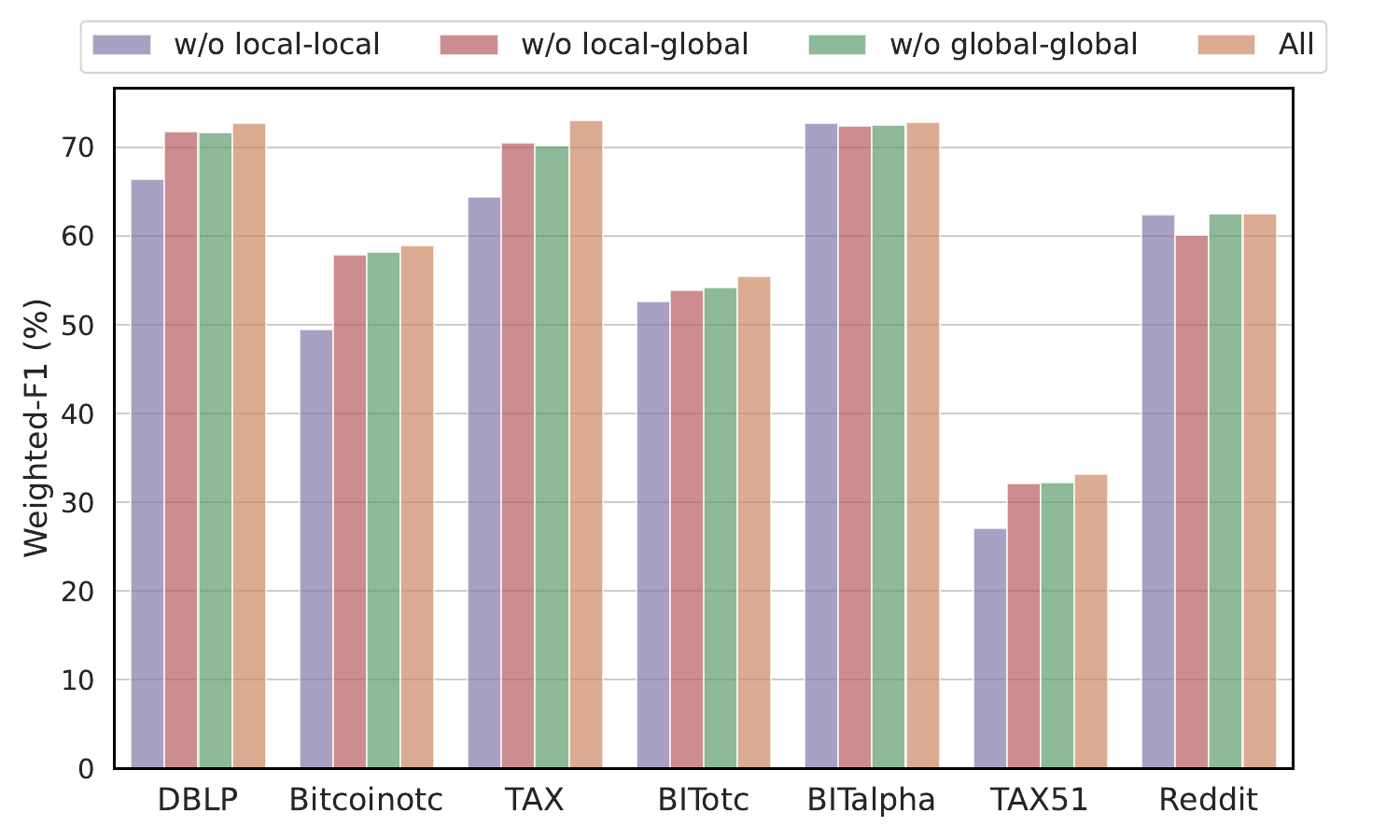}}\\
\caption{Ablation study on different variants.}
\label{fig:ablation}    
\end{figure*}

\begin{figure*}[!ht]
\centering
\subfloat[]{
\label{fig:param1}
\includegraphics[width=0.45\linewidth]{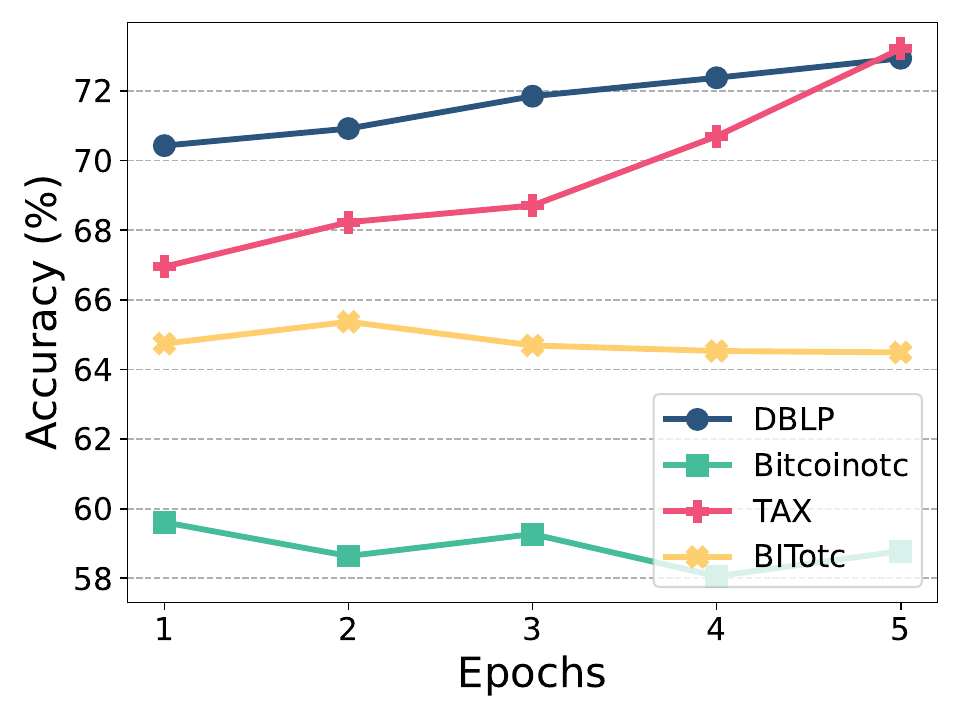}
}
\subfloat[]{
\label{fig:param2}
\includegraphics[width=0.45\linewidth]{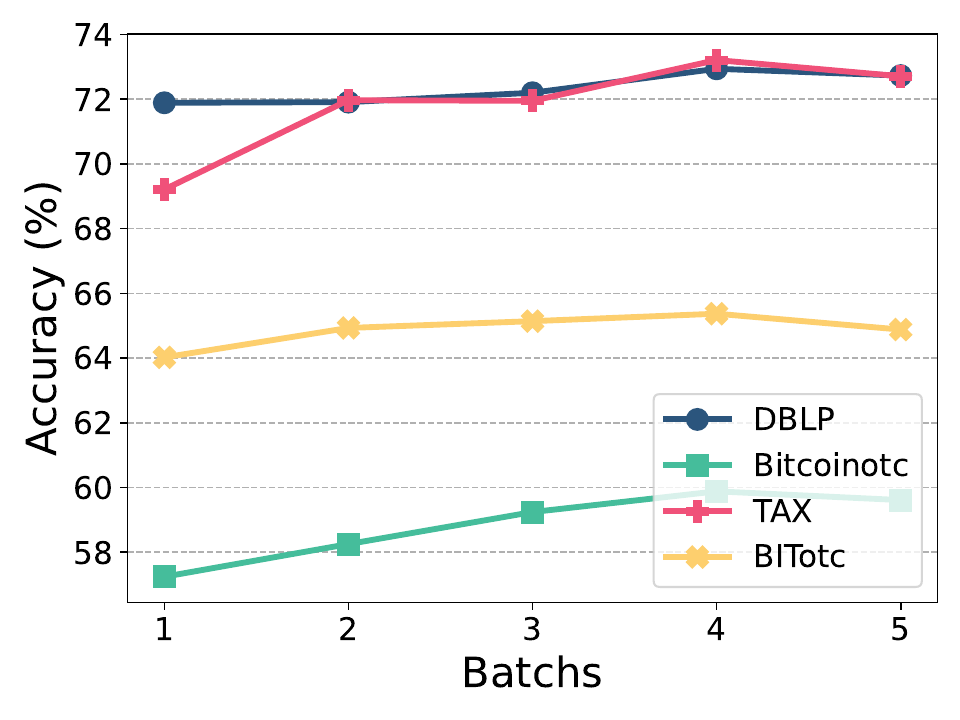}
}\\
\subfloat[]{
\label{fig:param3}
\includegraphics[width=0.45\linewidth]{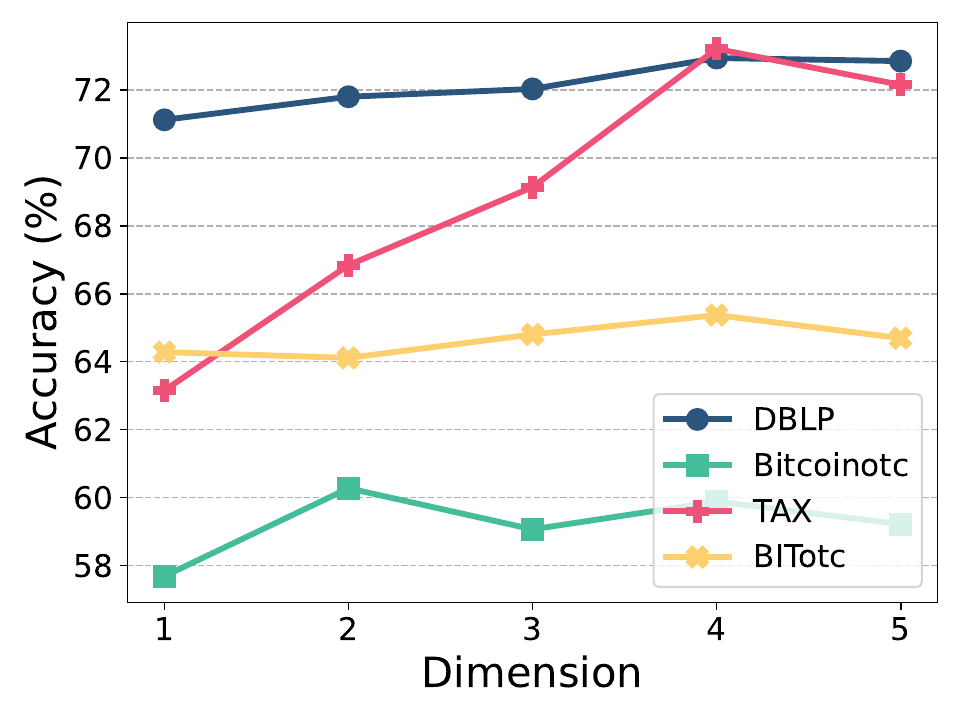}
}
\subfloat[]{
\label{fig:param4}
\includegraphics[width=0.45\linewidth]{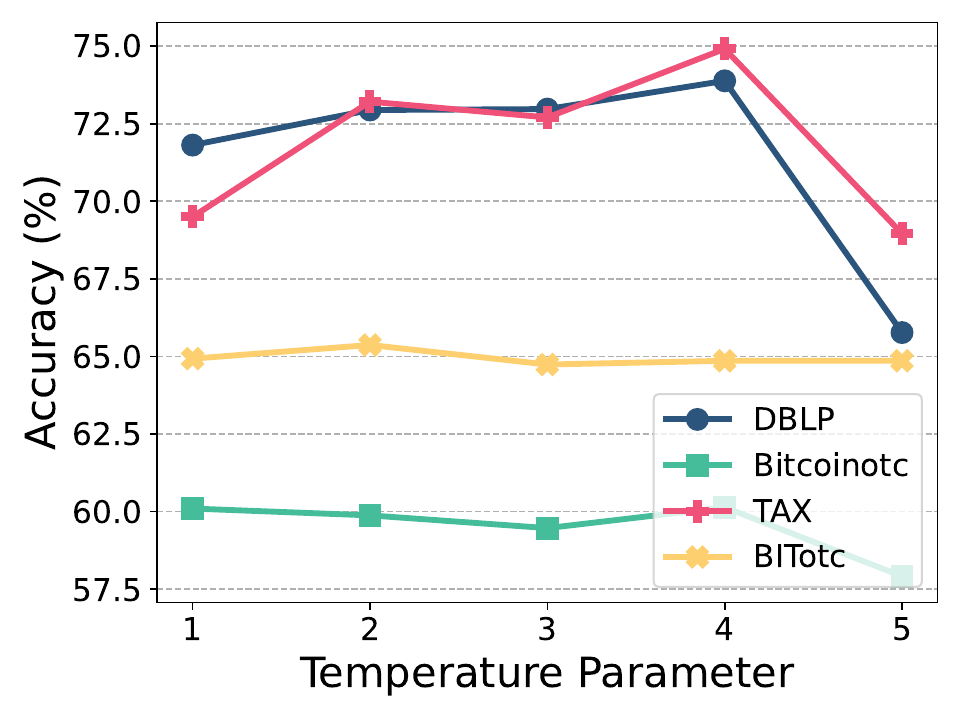}
}
\caption{Parameter sensitivity of CLDG++. Effect of  (a) epoch, (b) batch size, (c) hidden layer dimension, and (d) temperature parameter on the node classification.}
\label{fig:parameters}
\end{figure*}

\subsection{Ablation Study ($\mathbf{RQ_{6}}$)}
To validate the effectiveness of each contrast pair in CLDG++, we analyze the impact of local-local contrast loss, local-global contrast loss, and global-global contrast loss within our method. "W/o" denotes the removal of the loss during training. The results of our ablation study on 7 datasets are presented in Figure~\ref{fig:ablation}. It is evident that CLDG++ consistently outperforms other variants regarding overall performance. Specifically, the absence of local-local contrast, local-global contrast, and global-global contrast leads to performance degradation by 3.88\%, 0.92\% and 0.74\% in accuracy (\%), and 4.78\%, 1.41\% and 1.03\% in weighted-F1 (\%), respectively. This observation provides evidence that the local-local contrast plays a dominant role in representation learning. In summary, the performance degradation observed when any loss function is absent in CLDG++ highlights the importance of each contrast pair.

\subsection{Hyperparameters Sensitivity ($\mathbf{RQ_{7}}$)}
We investigate the sensitivity of the parameters and report the Accuracy (\%) results for four datasets with various parameters in Figure~\ref{fig:parameters}.

\textbf{Effect of epoch.}\quad We initially investigate the effect of epochs on the performance of CLDG++. The results, depicted in Fig.~\ref{fig:param1}, demonstrate that the performance of the DBLP and TAX datasets steadily improves as the number of epochs increases. Conversely, the Bitcoinotc and BITotc datasets exhibit minimal variance in performance as the number of epochs grows. This discrepancy can be attributed to the fact that larger datasets necessitate a greater number of epochs for training, whereas smaller datasets require fewer epochs to converge to the optimal solution.

\textbf{Effect of batch size.}\quad We then examine the effect of batch size on the performance of CLDG++. The corresponding results are presented in Fig.~\ref{fig:param2}. Across all four datasets, we observe a general improvement in classification performance as the batch size increases. However, when the batch size is 512, performance degradation is observed on some datasets. One possible reason is that a larger batch size causes more nodes with the same label to be pulled apart in the Euclidean space. Adjusting appropriate temperature parameters $\tau$ may alleviate this problem.

\textbf{Effect of hidden layer dimension.}\quad We explore the impact of output dimension on CLDG++. The results are shown in Fig.~\ref{fig:param3}. Notably, on the DBLP and TAX datasets, we observe an improvement in classification performance as the output dimensionality increases. However, on the Bitcoinotc and BITotc datasets, the classification performance initially improves but eventually diminishes with increasing output dimension. The reason is that larger hidden layer dimensions may introduce additional redundancy on smaller datasets.

\textbf{Effect of temperature parameter.}\quad We investigate the effect of varying the value of the temperature parameter $\tau$ in Eq.~\ref{eq:infonce}, which is often used to control the sharpness of the similarity scores or logits produced by the contrastive loss function. When the temperature coefficient is large, the similarity between samples becomes more difficult to distinguish. Specifically, on DBLP, Bitcoinotc, and TAX datasets, setting $\tau=0.5$ significantly reduces the model's accuracy and generalization ability. Conversely, when the temperature coefficient is small, excessive attention to marginal similarity differences can also lead to a performance decline.

\section{Conclusions}
To address the limitations of supervision and the disregard for temporal signals in graphs, we introduce a novel assumption known as temporal translation invariance. This assumption allows us to extend the unsupervised paradigm to dynamic graphs, enabling a more comprehensive understanding of the underlying patterns in the data.
Specifically, we proposed a sampling layer to extract the temporally-persistent signals. Then, CLDG encourages the node to maintain consistent local representations, i.e., temporal translation invariance under the timespan views. In addition, we also propose its variant CLDG++, which reveals the global context through diffusion matrices and utilizes three different views, local-local, local-global, and global-global, to compose multi-scale contrastive learning objectives to mine rich feature, structural, and temporal information in dynamic graphs. Finally, we introduce a new unsupervised anomaly detection framework that applies the concept of temporal translation invariance to dynamic graph anomaly detection tasks. The experiments demonstrate that CLDG and CLDG++ are superior state-of-the-art unsupervised methods and are competitive with supervised methods in node classification and anomaly detection tasks. Future work will focus on lightweight explicit modeling and capturing the evolutionary information in the graph to delve deeper into the rich information inherent in the dynamic graph.

\section*{Acknowledgment}
This research was partially supported by the National Natural Science Foundation of China, No.  (62050194, 62002282, 62192781, 61721002 and 62302380), the China Postdoctoral Science Foundation No. (2020M683492, 2023M742789), the MOE  Innovation Research Team No. (IRT\_17R86), Project of XJTU  Undergraduate Teaching Reform No. (20JX04Y), and Project of XJTU-SERVYOU  Joint Tax-AI Lab.








% \end{thebibliography}

\appendix
 \bibliographystyle{elsarticle-harv} 
 \bibliography{cas-refs}

\end{document}